\documentclass[a4paper,fleqn]{cas-sc}

\usepackage[numbers]{natbib}

\usepackage[ruled,linesnumbered]{algorithm2e}
\usepackage{calc}
\usepackage{subcaption}
\usepackage{pifont}
\usepackage{tikz}

\newtheorem{definition}{Definition}

\newcommand{\mat}[1]{\ensuremath{\boldsymbol{#1}}}
\newcommand{\tensor}[1]{\ensuremath{\boldsymbol{\mathcal{#1}}}}

\DeclareMathOperator{\hre}{HRE}

\begin{document}
\let\WriteBookmarks\relax
\def\floatpagepagefraction{1}
\def\textpagefraction{.001}

\shorttitle{Hierarchical rank-evolving representation for physics-informed neural networks}

\shortauthors{R. Su et~al.}

\title [mode = title]{Hierarchical rank-evolving representation for physics-informed neural networks}



%

\author[1]{Ruoyang Su}


\ead{202511110508@std.uestc.edu.cn}



\affiliation[1]{organization={University of Electronic Science and Technology of China},
            addressline={No.2006, Xiyuan Ave, West Hi-Tech Zone},
            city={Chengdu},
            postcode={611731},
            state={Sichuan},
            country={P.R.China}}

\author[1]{Xi-Le Zhao}[orcid=0000-0002-6540-946X]

\cormark[1]


\ead{xlzhao122003@163.com}

\ead[url]{https://zhaoxile.github.io/}



\author[2]{Kun Li}

\ead{likun@swufe.edu.cn}
\ead[url]{https://scholar.google.com/citations?user=7d_uncYAAAAJ&hl=zh-CN}

\affiliation[2]{organization={Southwestern University of Finance and Economics},
	addressline={555, Liutai Avenue, Wenjiang District},
	city={Chengdu},
	postcode={611130},
	state={Sichuan},
	country={P.R.China}}

\author[1]{Liang Li}[orcid=0000-0002-3847-0310]

\ead{plum.liliang@gmail.com}

\cortext[1]{Corresponding author}



\begin{abstract}
Recently, tensor-based physics-informed neural networks (T-PINNs) have received increasing attention. However, existing T-PINNs still face a fundamental challenge: they mainly rely on pre-specified low-rank tensor decompositions with manually tuned ranks, which limits their ability to capture the underlying structures of multivariate solution functions and hinders their practical deployment.
To address this challenge, we propose a hierarchical rank-evolving (abbreviated as HRE) representation for multivariate functions, which endows us to faithfully capture the underlying structure of the targeted multivariate function accompanying with automatic rank determination.
Concretely, in the hierarchical design of HRE representation, the target multivariate function is decomposed as a small-scale inner tensor with a set of univariate functions along each mode, where a customized tensor network decomposition can be readily deployed to capture the underlying structure of the small-scale inner tensor.
In HRE representation, the crucial hyperparameters, ranks, can be adaptively revealed during the decomposition, freeing us from manual rank tuning and making HRE practically applicable to real-world problems.
Besides, we build the HRE-PINNs correspondingly. Extensive numerical experiments, including high-dimensional static problems (Helmholtz equation and Poisson equation), nonlinear time-dependent problems (Klein-Gordon equation), and complex fluid-dynamics problems (flow mixing equation and Navier-Stokes equation), demonstrate that HRE-PINNs consistently outperform existing state-of-the-art approaches in terms of accuracy.
\end{abstract}




\maketitle

\section{Introduction}

Partial differential equations (PDEs) constitute the core mathematical tool for characterizing physical phenomena spanning numerous scientific and engineering fields, such as fluid dynamics \cite{song2025fenn,xing2025a}, solid mechanics \cite{rocha2023deepbnd,wurth2024physics}, electromagnetics \cite{mao2026accurate,rappaport2026a}, and quantum physics \cite{han2019solving,shang2025solving}. Obtaining accurate and efficient solutions to PDEs, especially high-dimensional and strongly nonlinear ones, has long been a central challenge in computational science. Traditional numerical methods, such as the finite difference method \cite{wang2012a}, finite element method \cite{martinezlera2024a,lu2026non}, and spectral methods \cite{li2022hermite}, have achieved great success in low-to-moderate dimensional problems, but they inherently suffer from the curse of dimensionality \cite{han2018solving,hutzenthaler2022overcoming,hu2024tackling}: the computational cost and memory footprint grow exponentially with the increase of the problem's dimension, making them computationally prohibitive for high-dimensional PDEs, such as those arising in many-body quantum mechanics, kinetic theory, and parametric uncertainty quantification.

Over the past few years, physics-informed neural networks (PINNs) \cite{raissi2019physics,cuomo2022scientific,xu2025on,cao2025adaptive,si2026complex} have emerged as a powerful mesh-free alternative for solving forward and inverse PDE problems. At their core, PINNs leverage the universal approximation capability of deep neural networks \cite{calin2020universal} to represent the solution function, and embed the governing physical laws, initial conditions, and boundary conditions directly into the loss function via automatic differentiation. This paradigm offers several transformative advantages over classical numerical methods: it eliminates the need for mesh generation \cite{pan2024domain,wang2025famaw}, naturally handles irregular geometries \cite{wang2023learning,sahlicostabal2024delta} with sparse or noisy observational data \cite{yang2021b,zou2024correcting}, and provides a continuous representation of the solution function across the entire spatio-temporal domain \cite{petersen2025pinnmep}. Thanks to these merits, PINNs have been successfully applied to a wide spectrum of PDE systems \cite{wang2024respecting,wang2024computing}, and have spurred a multitude of variants to address its practical limitations \cite{krishnapriyan2021characterizing,wang2021understanding,guo2023pre}.

Despite these advances, the curse of dimensionality remains a fundamental bottleneck that limits the application of standard PINNs to high-dimensional PDEs \cite{zhang2025annealed,lin2025monte,zhao2026casual}. The standard PINN uses a single multi-layer perceptron (MLP) that takes the $d$-dimensional coordinate vector as input and outputs the corresponding value of the solution function \cite{raissi2019physics}. For a $d$-dimensional domain discretized with $N$ points along each axis, the number of collocation points required for accurate residual evaluation scales as $O(N^d)$, leading to an exponential growth in the number of forward and backward passes through the network, as well as the computational cost of Jacobian and Hessian calculations via automatic differentiation \cite{cho2023separable,vemuri2025functional}. Even with modern GPU acceleration, standard PINNs struggle to handle PDEs with more than 4 dimensions, as the memory overhead and training time become intractable. Moreover, the single MLP architecture of standard PINNs fails to explicitly exploit the complex underlying structures in the solution functions of many physical PDEs, resulting in inefficient representation and slow convergence for high-dimensional problems.

To mitigate the curse of dimensionality in PINNs, a promising line of research has emerged that integrates the classical variable separation technique with low-rank tensor decompositions \cite{liu2022tt,cho2023separable,wang2024tensor,wang2024solving,vemuri2025functional,vemuri2026scalable}, extending discrete tensor algebra to the continuous function representation. The core idea of these methods is to factorize the high-dimensional solution function into a composition of univariate functions, thereby reducing the computational complexity from exponential to linear with respect to the problem dimension. For instance, the separable PINN framework \cite{cho2023separable} employs separate univariate MLPs for each coordinate axis, and constructs the solution via a function canonical-polyadic (CP) decomposition, achieving dramatic speedups for 3D and 4D PDEs by reducing the number of network forward passes from $O(N^d)$ to $O(Nd)$. Mandl et al. \cite{mandl2025separable} further extended this separable function decomposition paradigm to physics-informed deep neural operators (DeepONets), developing the Sep-PI-DeepONet framework that breaks the dimensionality barrier for operator learning in high-dimensional parametric PDEs. Beyond CP decomposition, Other related works \cite{vemuri2025functional} have also explored Tucker decomposition, tensor train (TT) decomposition and tensor ring (TR) decomposition for PINNs, further demonstrating the great potential of multivariate function decompositions in addressing the curse of dimensionality.
While these existing multivariate function decomposition methods have achieved notable success in reducing the computational complexity of PINNs, they suffer from critical limitations that hinder their performance on complex high-dimensional PDEs. Most existing works rely on primitive and fixed structures, such as CP decomposition, Tucker decomposition, TT decomposition, or TR decomposition, which fail to capture the complex structure inherent in the solution functions of many physical PDEs, leading to insufficient representation capability for high-precision approximation. Furthermore, these approaches lack an effective mechanism for automatically determining the suitable rank, which relies heavily on manual presetting and introduces extra hyperparameter tuning costs.

In this paper, we address these limitations by proposing a hierarchical rank-evolving (abbreviated as HRE) representation for multivariate functions, which endows us to faithfully capture the underlying structure of the targeted multivariate function accompanying with automatic rank determination.
Concretely, in the hierarchical design of HRE representation, the target multivariate function is decomposed as a small-scale inner tensor with a set of univariate functions along each mode, where a customized tensor network decomposition can be readily deployed to capture the underlying structure of the small-scale inner tensor. In HRE representation, the crucial hyperparameters, ranks, can be adaptively revealed during the decomposition, freeing us from manual rank tuning and making HRE practically applicable to real-world problems.
We build the HRE-PINNs correspondingly, and validate HRE-PINNs on a suite of benchmark PDEs, including high-dimensional static problems (3D Helmholtz equation and 5D Poisson equation), nonlinear time-dependent problems ((2+1)D Klein-Gordon equation), and complex fluid-dynamics problems ((2+1)D flow mixing equation and Navier-Stokes equation). Numerical results demonstrate that HRE-PINNs consistently outperform state-of-the-art PINNs and their variants in terms of accuracy.

The contributions of this paper are as follows:

1. We propose a hierarchical rank-evolving (abbreviated as HRE) representation for multivariate functions, which endows us to faithfully capture the underlying structure of the targeted multivariate function accompanying with automatic rank determination.

2. As crucial hyperparameters of HRE representation, ranks, can be adaptively revealed during the decomposition, freeing us from manual rank tuning and making HRE practically applicable to real-world problems.

3. We build the HRE-PINNs correspondingly, and conduct extensive numerical experiments on various benchmark PDEs. The results demonstrate that HRE-PINNs consistently outperform existing state-of-the-art approaches in terms of accuracy.

\section{Related work}

\subsection{Tensor network decomposition}

\begin{figure}[pos=!htb]
	\centering
	\includegraphics[scale=0.5]{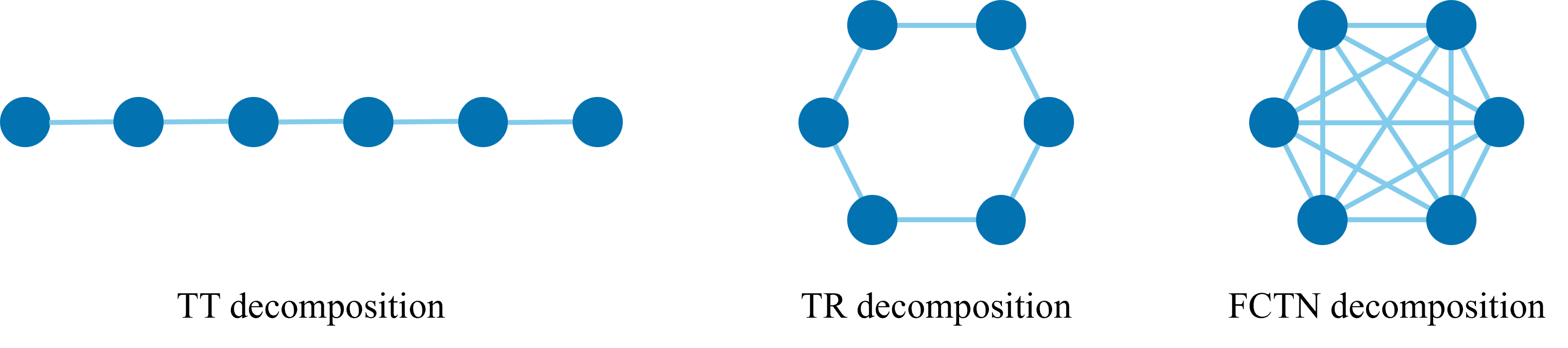}
	\caption{Underlying structures of tensor network decompositions.} \label{fig:tensor network decompositions}
\end{figure}

Recently, tensor network decompositions have emerged and shown great ability to process high-order tensors.
In tensor train (TT) decomposition \cite{oseledets2011tensor}, an $N$th-order tensor $\tensor{X} \in \mathbb{R}^{I_1 \times I_2 \times \cdots \times I_N}$ is decomposed into a matrix $\mat{U}_1 \in \mathbb{R}^{I_1 \times J_1}$, $N-2$ third-order tensors $\tensor{U}_n \in \mathbb{R}^{J_{n-1} \times I_n \times J_n}$ ($n=2,3,\cdots,N-1$) and another matrix $\mat{U}_N \in \mathbb{R}^{J_{N-1} \times I_N}$ multiplied sequentially:
\[ \tensor{X}(i_1, i_2, i_3, \cdots i_N) = \sum_{j_1=1}^{J_1} \sum_{j_2=1}^{J_2} \sum_{j_3=1}^{J_3} \cdots \sum_{j_{N-1}=1}^{J_{N-1}} \mat{U}_1(i_1, j_1) \tensor{U}_2(j_1, i_2, j_2) \tensor{U}_3(j_2, i_3, j_3) \cdots \mat{U}_N(j_{N-1}, i_N). \]
In tensor ring (TR) decomposition \cite{zhao2016tensor}, an $N$th-order tensor $\tensor{X} \in \mathbb{R}^{I_1 \times I_2 \times \cdots \times I_N}$ is decomposed into $N$ third-order tensors $\tensor{U}_n \in \mathbb{R}^{J_{n-1} \times I_n \times J_n}$ ($n=1,2,\cdots,N$, $J_0=J_N$) multiplied circularly:
\[ \tensor{X}(i_1, i_2, \cdots i_N) = \sum_{j_0=1}^{J_0} \sum_{j_1=1}^{J_1} \cdots \sum_{j_{N-1}=1}^{J_{N-1}} \tensor{U}_1(j_0, i_1, j_1) \tensor{U}_2(j_1, i_2, j_2) \cdots \tensor{U}_N(j_{N-1}, i_N, j_0). \]
TT decomposition and TR decomposition only establish operations between adjacent two factors and are highly sensitive to the permutation of tensor modes, resulting in an inadequate and inflexible representation.
To address the limitations of TT decomposition and TR decomposition, fully-connected tensor network (FCTN) decomposition \cite{zheng2021fully,zheng2022tensor} was proposed, in which an $N$th-order tensor $\tensor{X} \in \mathbb{R}^{I_1 \times I_2 \times \cdots \times I_N}$ is decomposed into a set of $N$th-order factors $\tensor{U}_n \in \mathbb{R}^{J_{1,n} \times J_{2,n} \times \cdots \times J_{n-1,n} \times I_n \times J_{n,n+1} \times \cdots \times J_{n,N}}$ with multi-linear operations between any two factors:
\begin{align*}
	\tensor{X}(i_1, i_2, \cdots, i_N) & = \sum_{j_{1,2}}^{J_{1,2}} \sum_{j_{1,3}}^{J_{1,3}} \cdots \sum_{j_{1,N}}^{J_{1,N}} \sum_{j_{2,3}}^{J_{2,3}} \cdots \sum_{j_{2,N}}^{J_{2,N}} \cdots \sum_{j_{N-1,N}}^{J_{N-1,N}} \\
	& \mathrel{\phantom{=}} \tensor{U}_1(i_1, j_{1,2}, j_{1,3}, \cdots, j_{1,N}) \tensor{U}_2(j_{1,2}, i_2, j_{2,3}, \cdots, j_{2,N}) \cdots \\
	& \mathrel{\phantom{=}} \tensor{U}_n(j_{1,n}, j_{2,n}, \cdots, j_{n-1,n}, i_n, j_{n,n+1}, \cdots, j_{n,N}) \cdots \tensor{U}_N(j_{1,N}, j_{2,N}, \cdots, j_{N-1,N}, i_N).
\end{align*}
One primary advantage of FCTN decomposition is its ability to directly capture the intrinsic all-mode correlations of the target tensor, while TT decomposition and TR decomposition only establish the connections between adjacent two factors.
Figure \ref{fig:tensor network decompositions} illustrates the underlying structures of the aforementioned tensor network decompositions.

However, the underlying structures of the aforementioned tensor network decompositions are all fixed, and finding the optimal structure has always been a challenging problem. To address this issue, SVD-inspired tensor network (SVDinsTN) decomposition \cite{zheng2024svdinstn} was proposed to efficiently search for a customized structure to achieve a compact representation. By inserting a diagonal matrix for each edge of the fully-connected tensor network, SVDinsTN can optimize tensor network factors and diagonal matrices simultaneously, evolving a compact tensor network structure.

\subsection{Physics-informed neural network}

Suppose that $\varOmega \subseteq \mathbb{R}^d$ is a spatial domain with boundary $\partial \varOmega$. Consider the general formulation of PDEs:
\begin{equation} \label{eq:general form of pdes}
	\begin{cases}
		\mathscr{L}u(\mat{x}) = f(\mat{x}), & \mat{x} \in \varOmega, \\
		\mathscr{B}u(\mat{x}) = b(\mat{x}), & \mat{x} \in \partial \varOmega,
	\end{cases}
\end{equation}
where $\mathscr{L}$ denotes the partial differential operator, $\mathscr{B}$ denotes the boundary operator, $f(\mat{x})$ is the source term, and $b(\mat{x})$ is the boundary condition.

Physics-informed neural networks (PINNs) are a class of neural networks that embed the physical constraints governed by PDEs directly into the loss function, thereby biasing the optimization process toward physically consistent solutions. PINNs aim to approximate the unknown solution function $u(\mat{x})$ in \eqref{eq:general form of pdes} using a parameterized neural network $\tilde{u}_\theta(\mat{x})$, where $\theta$ denotes the set of trainable parameters. The total loss function is constructed as
\[ L(\theta) = L_\mathrm{R}(\theta) + \lambda L_\mathrm{B}(\theta), \]
where the residual loss
\[ L_\mathrm{R}(\theta) = \frac{1}{N_\mathrm{R}} \sum_{m=1}^{N_\mathrm{R}} \left| \mathscr{L}\tilde{u}_\theta(\mat{x}_\mathrm{R}^{(m)}) - f(\mat{x}_\mathrm{R}^{(m)}) \right|^2, \qquad \left\{\mat{x}_\mathrm{B}^{(m)}\right\}_{m=1}^{N_\mathrm{R}} \subseteq \varOmega \]
penalizes violations of the governing equation, and the boundary loss
\[ L_\mathrm{B}(\theta) = \frac{1}{N_\mathrm{B}} \sum_{m=1}^{N_\mathrm{B}} \left| \mathscr{B}\tilde{u}_\theta(\mat{x}_\mathrm{B}^{(m)}) - b(\mat{x}_\mathrm{B}^{(m)}) \right|^2, \qquad \left\{\mat{x}_\mathrm{B}^{(m)}\right\}_{m=1}^{N_\mathrm{B}} \subseteq \partial \varOmega \]
enforces satisfaction of the boundary condition.

By introducing physical priors into the learning objective, PINNs exhibit several distinctive advantages: they require no labeled data for training, enabling purely physics-constrained solution of PDEs \cite{li2026an}; they avoid the meshing requirements of traditional numerical methods, making them highly flexible for irregular and high-dimensional domains \cite{kashefi2022physics,hu2025bias}; and they seamlessly integrate data and physics, offering a robust paradigm for forward and inverse problems in scientific computing \cite{arzani2021uncovering}.

\section{Methodology}

\subsection{Notations and preliminaries}

In this paper, we follow the notational convention where $a$, $\mat{a}$, $\mat{A}$, and $\tensor{A}$ stand for scalars, vectors, matrices and tensors correspondingly. For an $N$th‑order tensor $\tensor{A} \in \mathbb{R}^{I_1 \times I_2 \times \cdots \times I_N}$, $\tensor{A}(i_1, i_2, \cdots, i_N)$ gives its $(i_1, i_2, \cdots, i_N)$th element.

\begin{definition}[Function Tucker decomposition \cite{vemuri2025functional,luo2024low}]
	For a multivariate function $f \colon \mathbb{R}^N \to \mathbb{R}$, the function Tucker decomposition is defined as
	\[ f(x_1, x_2, \cdots, x_N) = \sum_{i_1=1}^{I_1} \sum_{i_2=1}^{I_2} \cdots \sum_{i_N=1}^{I_N} \tensor{A}(i_1, i_2, \cdots, i_N) \cdot \mat{g}_1(x_1)(i_1) \cdot \mat{g}_2(x_2)(i_2) \cdot \cdots \cdot \mat{g}_N(x_N)(i_N), \]
	where $\tensor{A} \in \mathbb{R}^{I_1 \times I_2 \times \cdots \times I_N}$ is the core tensor, $\mat{g}_n \colon \mathbb{R} \to \mathbb{R}^{I_n}$ is a vector-valued function, and $\mat{g}_n(x_n)(i_n) \in \mathbb{R}$ denotes the $n$th element of the vector $\mat{g}_n(x_n) \in \mathbb{R}^{I_n}$ ($n=1,2,\cdots,N$).
\end{definition}

\subsection{Hierarchical rank-evolving representation for multivariate functions}

Tensor-based PINNs have been widely adopted to mitigate the curse of dimensionality. Typical approaches include function canonical-polyadic (CP) decomposition, Tucker decomposition, tensor train (TT) decomposition, and tensor ring (TR) decomposition \cite{cho2023separable, vemuri2025functional}. These methods factorize a high-dimensional solution function into a combination of univariate functions, reducing the computational complexity from exponential to linear. However, they rely on primitive low-rank tensor decompositions with fixed underlying structures, which are often insufficient to capture the complex correlations inherent in the solution functions of many physical PDEs. Moreover, these approaches lack an automatic mechanism for determining the optimal ranks, requiring manual presetting that introduces extra hyperparameter tuning costs.

To address these limitations, we propose a hierarchical rank-evolving (abbreviated as HRE) representation for multivariate functions, which endows us to faithfully capture the underlying structure of the targeted multivariate function accompanying with automatic rank determination. Concretely, in the hierarchical design of HRE representation, the target multivariate function is decomposed as a small-scale inner tensor with a set of univariate functions along each mode, where a customized tensor network decomposition can be readily deployed to capture the underlying structure of the small-scale inner tensor.

\begin{definition}[Hierarchical rank-evolving representation]
	For a multivariate function $f \colon \mathbb{R}^N \to \mathbb{R}$, the hierarchical rank-evolving (abbreviated as HRE) representation is defined as
	\begin{align*}
		f(x_1, x_2, \cdots, x_N) = \sum_{i_1=1}^{I_1} \sum_{i_2=1}^{I_2} \cdots \sum_{i_N=1}^{I_N} &\ \tensor{A}(i_1, i_2, \cdots, i_N) \cdot \mat{r}^{(\mathrm{out})}_1(i_1) \cdot \mat{g}_1(x_1)(i_1) \cdot \mat{r}^{(\mathrm{out})}_2(i_2) \cdot \mat{g}_2(x_2)(i_2) \\
		& \cdot \cdots \cdot \mat{r}^{(\mathrm{out})}_N(i_N) \cdot \mat{g}_N(x_N)(i_N),
	\end{align*}
	where the core tensor
	\begin{align*}
		\tensor{A}(i_1, i_2, \cdots, i_N) & = \sum_{j_{1,2}}^{J_{1,2}} \sum_{j_{1,3}}^{J_{1,3}} \cdots \sum_{j_{1,N}}^{J_{1,N}} \sum_{j_{2,3}}^{J_{2,3}} \cdots \sum_{j_{2,N}}^{J_{2,N}} \cdots \sum_{j_{N-1,N}}^{J_{N-1,N}} \\
		& \mathrel{\phantom{=}} \mat{r}^{(\mathrm{in})}_{1,2}(j_{1,2}) \mat{r}^{(\mathrm{in})}_{1,3}(j_{1,3}) \cdots \mat{r}^{(\mathrm{in})}_{1,N}(j_{1,N}) \mat{r}^{(\mathrm{in})}_{2,3}(j_{2,3}) \cdots \mat{r}^{(\mathrm{in})}_{2,N}(j_{2,N}) \cdots \mat{r}^{(\mathrm{in})}_{N-1,N}(j_{N-1,N}) \\
		& \mathrel{\phantom{=}} \tensor{B}_1(i_1, j_{1,2}, j_{1,3}, \cdots, j_{1,N}) \tensor{B}_2(j_{1,2}, i_2, j_{2,3}, \cdots, j_{2,N}) \cdots \\
		& \mathrel{\phantom{=}} \tensor{B}_n(j_{1,n}, j_{2,n}, \cdots, j_{n-1,n}, i_n, j_{n,n+1}, \cdots, j_{n,N}) \cdots \tensor{B}_N(j_{1,N}, j_{2,N}, \cdots, j_{N-1,N}, i_N),
	\end{align*}
	Here, $\mat{g}_n \colon \mathbb{R} \to \mathbb{R}^{I_n}$ is a vector-valued function, $\mat{g}_n(x_n)(i_n) \in \mathbb{R}$ denotes the $n$th element of the vector $\mat{g}_n(x_n) \in \mathbb{R}^{I_n}$, $\mat{r}^{(\mathrm{out})}_n \in \mathbb{R}^{I_n}$, $\tensor{B}_n \in \mathbb{R}^{J_{1,n} \times J_{2,n} \times \cdots \times J_{n-1,n} \times I_n \times J_{n,n+1} \times \cdots \times J_{n,N}}$ ($n=1,2,\cdots,N$), and $\mat{r}^{(\mathrm{in})}_{k,l} \in \mathbb{R}^{J_{k,l}}$ ($1 \leqslant k \leqslant l \leqslant N$). For simplicity, we denote this decomposition as
	\[ f(x_1, x_2, \cdots, x_N) = \hre\left(\{\tensor{B}_n\}_{n=1}^N,\ \{\mat{g}_n(x_n)\}_{n=1}^N,\ \{\mat{r}^{(\mathrm{in})}_{k,l}\}_{1 \leqslant k \leqslant l \leqslant N},\ \{\mat{r}^{(\mathrm{out})}_n\}_{n=1}^N \right). \]
\end{definition}

The outer representation provides a variable-separable form: for any input $(x_1,x_2,\cdots,x_N)$, the function value is obtained by evaluating each $\mat{g}_n(x_n)$ and then contracting with the core tensor $\tensor{A}$. This reduces the evaluation cost from exponential to linear. The inner customized tensor network decomposition further capture arbitrary all-mode correlations through its flexible structure. More importantly, the learnable rank-evolving vectors $\{\mat{r}^{(\mathrm{in})}_{k,l}\}_{1 \leqslant k \leqslant l \leqslant N}$ (controlling the inner ranks) and $\{\mat{r}^{(\mathrm{out})}_n\}_{n=1}^N$ (controlling the outer ranks) are designed to be sparse. By incorporating $\ell^1$ regularization on these rank-evolving vectors into the loss function, we can effectively prune redundant edges. This process automatically determines the suitable ranks as well as the compact underlying structure of the inner tensor network, eliminating manual hyperparameter tuning.

The rank‑evolving capability of HRE representation stems from the learnable rank-evolving vectors $\{\mat{r}^{(\mathrm{in})}_{k,l}\}_{1 \leqslant k \leqslant l \leqslant N}$ and $\{\mat{r}^{(\mathrm{out})}_n\}_{n=1}^N$. In the outer representation, each $\mat{r}^{(\mathrm{out})}_n$ multiplies the output of the univariate network $\mat{g}_n(x_n)$ before contraction with the core tensor $\tensor{A}$. Thus, if an element of $\mat{r}^{(\mathrm{out})}_n$ becomes zero, the corresponding component $\mat{g}_n(x_n)(i_n)$ is equivalently removed, reducing the outer rank. Similarly, in the inner representation, each vector $\mat{r}^{(\mathrm{in})}_{k,l} \in \mathbb{R}_{J_{k,l}}$ modulates the connection between the $k$th and $l$th factors. When an element of $\mat{r}^{(\mathrm{in})}_{k,l}$ shrinks to zero, the corresponding rank is reduced. If a rank-evolving vector $\mat{r}^{(\mathrm{in})}_{k,l}$ becomes $\mat{0}$, the corresponding edge is pruned, which simplifies the underling structure.

The HRE representation thus unifies the expressive power of flexible tensor networks decomposition with rank-evolving mechanism, offering a powerful and adaptive tool for representing high-dimensional functions. When employed in PINNs to approximate the solution function of a PDE, we replace the standard neural network with HRE representation, and train all parameters jointly under the physics-informed loss augmented with sparsity regularization. The details are described next.

\subsection{Hierarchical rank-evolving representation-based physics-informed neural networks}

\begin{figure}[pos=!tb]
	\centering
	\includegraphics[width=\linewidth]{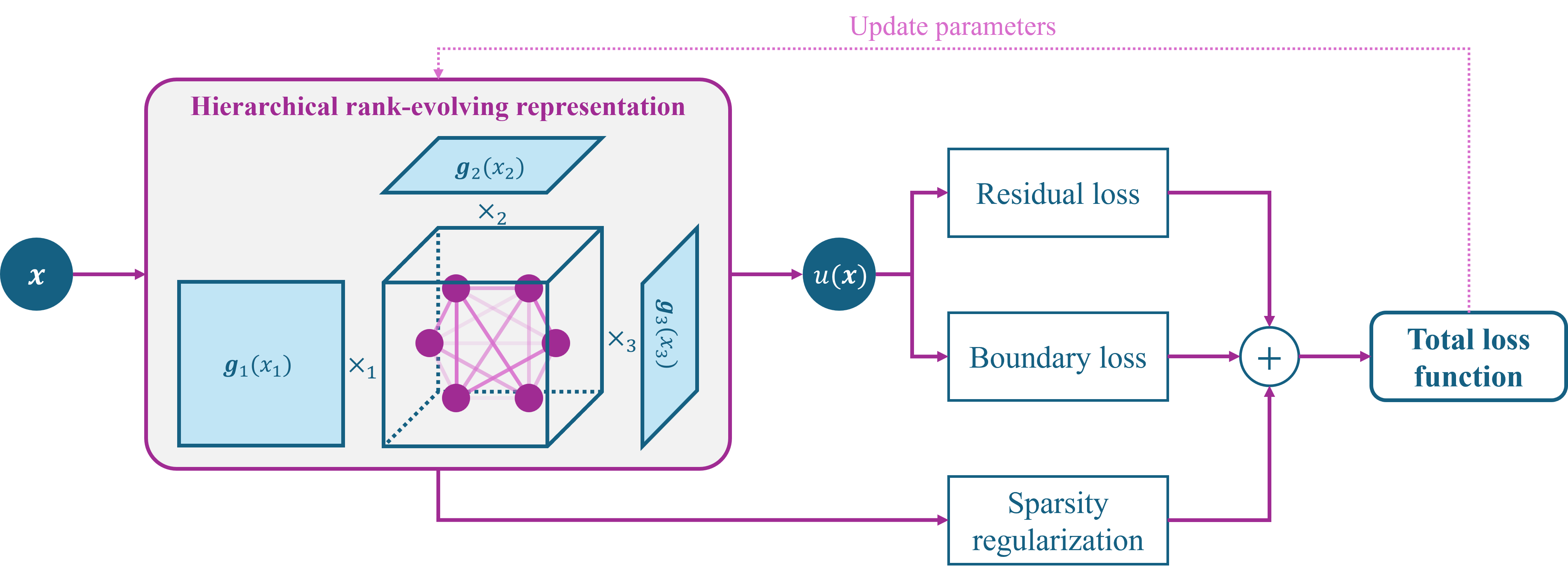}
	\caption{Flowchart of HRE-PINNs. In the hierarchical design of HRE representation, the target multivariate function is decomposed as a small-scale inner tensor with a set of univariate functions along each mode, where a customized tensor network decomposition can be readily deployed to capture the underlying structure of the small-scale inner tensor. We build the HRE-PINNs correspondingly, with a hybrid loss function that incorporates three components: the PDE residual loss, the boundary condition loss, and a sparsity regularization term on the rank-evolving vectors.} \label{fig:hre-pinn}
\end{figure}

We employ the hierarchical rank-evolving (HRE) representation to approximate the solution function of the general partial differential equation system given in \eqref{eq:general form of pdes}, and build the hierarchical rank-evolving representation-based physics-informed neural networks (abbreviated as HRE-PINNs).
In this framework, the unknown solution function $u(\mat{x})$ is approximated by
\begin{equation} \label{eq:forward of HRE-PINN}
	\tilde{u}(\mat{x}) = \hre\left(\{\tensor{B}_n\}_{n=1}^N,\ \{\mat{g}_n(x_n)\}_{n=1}^N,\ \{\mat{r}^{(\mathrm{in})}_{k,l}\}_{1 \leqslant k \leqslant l \leqslant N},\ \{\mat{r}^{(\mathrm{out})}_n\}_{n=1}^N \right), \qquad \mat{x} = (x_1, x_2, \cdots, x_N) \in \mathbb{R}^N,
\end{equation}
where each univariate vector-valued function $\mat{g}_n \colon \mathbb{R} \to \mathbb{R}^{I_n}$ is implemented by a neural network with parameters $\theta_n$.
Figure \ref{fig:hre-pinn} illustrates the flowchart of HRE-PINNs.

To automatically determine the suitable ranks and underlying structure, we augment the physics‑informed loss with $\ell^1$‑norm regularization on both $\{\mat{r}^{(\mathrm{in})}_{k,l}\}_{1 \leqslant k \leqslant l \leqslant N}$ and $\{\mat{r}^{(\mathrm{out})}_n\}_{n=1}^N$. During training, the regularization drives redundant components toward zero, while the PDE residual and boundary losses ensure that only those components essential for accurate solution approximation survive. As a result, the effective ranks are not predetermined but emerge from the optimization process: the number of non‑zero entries in each $\mat{r}^{(\mathrm{in})}_{k,l}$ and $\mat{r}^{(\mathrm{out})}_n$ after training directly gives the suitable rank. This mechanism eliminates the need for manual rank tuning, prevents over‑ or under‑fitting, and reveal the underlying structure of the target solution, which is a key advantage over existing tensor‑based PINNs with fixed underlying structure. Moreover, the sparsity of $\mat{r}^{(\mathrm{in})}_{k,l}$ simultaneously reveals a compact inner tensor network structure that best captures the all‑mode correlations of the core tensor.
Compared with previous tensor-based PINNs, the rank-evolving mechanism eliminates manual tuning of ranks and topological structures, improving both accuracy and training stability for high-dimensional PDEs.

In practice, most high-dimensional physical systems exhibit inherent low-rank and compact underlying structures, which render the rank-evolving vectors $\{\mat{r}^{(\mathrm{in})}_{k,l}\}_{1 \leqslant k \leqslant l \leqslant N}$ and $\{\mat{r}^{(\mathrm{out})}_n\}_{n=1}^N$ naturally sparse. Accordingly, we construct a hybrid loss function that incorporates three components: the PDE residual loss, the boundary condition loss, and a sparsity regularization term on the rank-evolving vectors:
\begin{equation} \label{eq:loss of HRE-PINN}
	\begin{split}
		& \mathrel{\phantom{=}} L\left( \{\tensor{B}_n\}_{n=1}^N,\ \{\theta_n\}_{n=1}^N,\ \{\mat{r}^{(\mathrm{in})}_{k,l}\}_{1 \leqslant k \leqslant l \leqslant N},\ \{\mat{r}^{(\mathrm{out})}_n\}_{n=1}^N \right) \\
		& = \frac{1}{N_\mathrm{R}} \sum_{m=1}^{N_\mathrm{R}} \left| \mathscr{L}\tilde{u}(\mat{x}_\mathrm{R}^{(m)}) - f(\mat{x}_\mathrm{R}^{(m)}) \right|^2 + \eta \frac{1}{N_\mathrm{B}} \sum_{m=1}^{N_\mathrm{B}} \left| \mathscr{B}\tilde{u}(\mat{x}_\mathrm{B}^{(m)}) - b(\mat{x}_\mathrm{B}^{(m)}) \right|^2 + \lambda \sum_{1 \leqslant k \leqslant l \leqslant N} \|\mat{r}^{(\mathrm{in})}_{k,l}\|_1 + \mu \sum_{n=1}^{N} \|\mat{r}^{(\mathrm{out})}_n\|_1,
	\end{split}
\end{equation}
where $\left\{\mat{x}_\mathrm{R}^{(m)}\right\}_{m=1}^{N_\mathrm{R}} \subseteq \varOmega$, $\left\{\mat{x}_\mathrm{B}^{(m)}\right\}_{m=1}^{N_\mathrm{B}} \subseteq \partial \varOmega$ are collocation point sets inside the domain $\varOmega$ and on the boundary $\partial \varOmega$, respectively.
The first term penalizes violations of the governing PDE, the second term enforces the satisfaction of boundary and initial conditions, and the third and fourth regularization terms promote the sparsity of rank-evolving vectors $\{\mat{r}^{(\mathrm{in})}_{k,l}\}_{1 \leqslant k \leqslant l \leqslant N}$ and $\{\mat{r}^{(\mathrm{out})}_n\}_{n=1}^N$, which facilitates adaptive discovery of the suitable rank and the compact topological structure. All trainable parameters in the overall framework, including $\{\tensor{B}_n\}_{n=1}^N$, $\{\theta_n\}_{n=1}^N$, $\{\mat{r}^{(\mathrm{in})}_{k,l}\}_{1 \leqslant k \leqslant l \leqslant N}$, and $\{\mat{r}^{(\mathrm{out})}_n\}_{n=1}^N$, are jointly optimized via the Adam optimizer \cite{kingma2015adam}.

\begin{algorithm}[tb]
	\caption{Hierarchical rank-evolving representation-based physics-informed neural networks}
	\KwIn{Target PDE $\mathscr{L} u = f$, boundary condition $\mathscr{B} u = b$, collocation points $\left\{\mat{x}_\mathrm{R}^{(m)}\right\}_{m=1}^{N_\mathrm{R}} \subseteq \varOmega$ and $\left\{\mat{x}_\mathrm{B}^{(m)}\right\}_{m=1}^{N_\mathrm{B}} \subseteq \partial \varOmega$, hyper-parameters $\lambda$ and $\mu$, maximum epochs number $M_\mathrm{e}$.}
	\KwOut{Predicted solution function $\tilde{u}$.}

	Initialize inner tensor network decomposition factors $\{\tensor{B}_n\}_{n=1}^N$, rank-evolving vectors $\{\mat{r}^{(\mathrm{in})}_{k,l}\}_{1 \leqslant k \leqslant l \leqslant N}$ and $\{\mat{r}^{(\mathrm{out})}_n\}_{n=1}^N$, and univariate neural networks $\{\mat{g}_n\}_{n=1}^N$ with trainable parameters $\{\theta_n\}_{n=1}^N$\;
	\For{$i = 1 : M_\mathrm{e}$}{
		Compute the HRE representation $\tilde{u}(\mat{x})$ for all collocation points via \eqref{eq:forward of HRE-PINN}\;
		Compute the regularization term $\sum_{1 \leqslant k \leqslant l \leqslant N} \|\mat{r}^{(\mathrm{in})}_{k,l}\|_1$ and $\sum_{n=1}^{N} \|\mat{r}^{(\mathrm{out})}_n\|_1$\;
		Calculate the total loss $L$ following \eqref{eq:loss of HRE-PINN}\;
		Update $\{\tensor{B}_n\}_{n=1}^N$, $\{\mat{r}^{(\mathrm{out})}_n\}_{n=1}^N$, $\{\mat{r}^{(\mathrm{in})}_{k,l}\}_{1 \leqslant k \leqslant l \leqslant N}$ and $\{\theta_n\}_{n=1}^N$ using the Adam optimizer \cite{kingma2015adam}\;
	}
	Generate the predicted solution function $\tilde{u}$ with the optimal parameters\;
\end{algorithm}

\section{Experiments}

To comprehensively evaluate the HRE-PINN, we select several representative PDEs spanning three categories: high-dimensional static problems (3D Helmholtz equation and 5D Poisson equation), nonlinear time-dependent problems ((2+1)D Klein-Gordon equation), and complex fluid-dynamics problems ((2+1)D flow mixing equation and (2+1)D Navier-Stokes equation).
All experiments are implemented in JAX with automatic differentiation for PDE residual calculation. The performance is evaluated by three metrics: the root mean square error, relative $L^2$ error and relative $L^\infty$ error between the exact solution and the predicted solution.

To comprehensively assess the approximation capability and generalization performance of our proposed HRE‑PINN framework, we carry out a series of numerical tests covering diverse PDE scenarios with increasing complexity. The test suite encompasses high‑dimensional stationary elliptic PDEs, nonlinear time‑dependent hyperbolic systems, as well as challenging fluid‑dynamics benchmarks featuring sharp interfaces and turbulent vortex interactions. These test cases are ordered by their inherent difficulty level, so that we can systematically reveal the advantages brought by the hierarchical rank‑evolving representation across different physical settings.

We compare HRE‑PINN against several state‑of‑the‑art tensor‑based PINN baselines, including SPINN, TT‑PINN and Tucker‑PINN. For fair comparison, all competing approaches are configured following the original publications. These baseline methods are implemented under the JAX‑based experimental environment, keeping consistent network capacities, collocation point budgets and optimizer settings. The whole computational pipeline relies on JAX's built‑in automatic differentiation module to compute PDE residual terms required for physics‑informed training.

To achieve objective assessment, three metrics are adopted to measure the discrepancy between predicted fields and reference solutions: root mean square error (RMSE), relative $L^2$ error and relative $L^\infty$ error. Lower values of these metrics correspond to more accurate PDE solution approximations. During model training, we monitor these metrics every $1\,000$ epochs. The model checkpoint yielding the best prediction accuracy on test samples is retained to produce the final results.

In the following subsections, we report numerical results for each benchmark sequentially: 3D Helmholtz equation, 5D Poisson equation, (2+1)D Klein‑Gordon equation, (2+1)D flow‑mixing equation, and (2+1)D Navier‑Stokes equation.

\subsection{3D Helmholtz equation}

Helmholtz equations are fundamental elliptic PDEs governing time-harmonic wave propagation in acoustics, electromagnetics, and structural dynamics. They are widely recognized as a challenging benchmark for physics-informed neural networks due to their high-frequency solution components, which frequently lead to training degradation and catastrophic accuracy loss in standard PINN formulations. We consider the 3D Helmholtz equation with homogeneous Dirichlet boundary conditions:
\[ \begin{cases}
	\Delta u(\mat{x}) + k^2 u(\mat{x}) = q(\mat{x}), & \mat{x} \in \varOmega, \\
	u(\mat{x}) = 0, & \mat{x} \in \partial\varOmega,
\end{cases} \]
where $\varOmega = [-1,1]^3$, and $k = 1$. We employ the manufactured solution to generate an exact reference solution:
\[ u(\mat{x}) = \sin(4 \pi x_1)\sin(4 \pi x_2)\sin(3 \pi x_3), \qquad \mat{x} = (x_1, x_2, x_3), \]
with the corresponding source term derived as
\[ q(\mat{x}) = - (4\pi)^2 u(\mat{x}) - (4\pi)^2 u(\mat{x}) - (3\pi)^2 u(\mat{x}) + k^2 u(\mat{x}). \]
This benchmark tests the ability of HRE-PINN to handle high-frequency functions.

\begin{table}[width=.7\linewidth,cols=3,pos=!htb]
	\caption{Performance comparison of different methods for 3D Helmholtz equation} \label{tab:helmholtz}
	\begin{tabular*}{\tblwidth}{@{}LLLL@{}}
		\toprule
		Method & RMSE & $L^2$ error & $L^\infty$ error \\
		\midrule
		SPINN & 0.014\,352 & 0.041\,210 & 0.136\,670 \\
		TT-PINN & 0.009\,998 & 0.028\,709 & 0.090\,714 \\
		Tucker-PINN & 0.012\,661 & 0.036\,356 & 0.104\,973 \\
		HRE-PINN & 0.005\,021 & 0.014\,417 & 0.054\,701 \\
		\bottomrule
	\end{tabular*}
\end{table}

\begin{figure}[pos=!htb]
	\centering
	\begin{tabular}{@{} @{\extracolsep{\fill}} m{1em} *{4}{>{\centering\arraybackslash}m{0.2\linewidth}} @{}}
		& SPINN & TT-PINN & Tucker-PINN & Proposed \\
		\rotatebox{90}{Predicted} &
		\includegraphics[width=\linewidth]{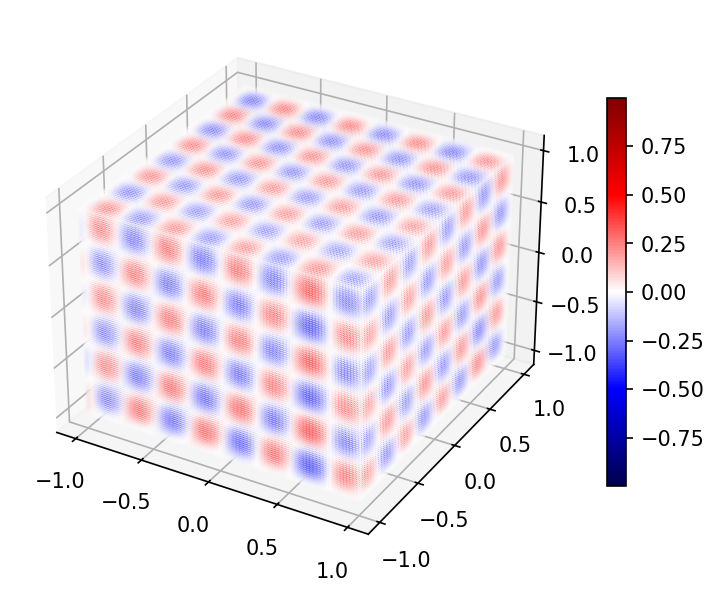} &
		\includegraphics[width=\linewidth]{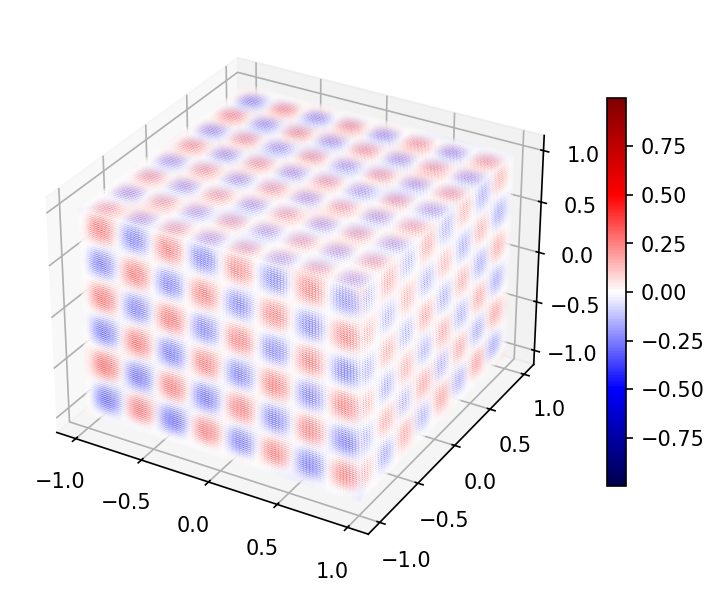} &
		\includegraphics[width=\linewidth]{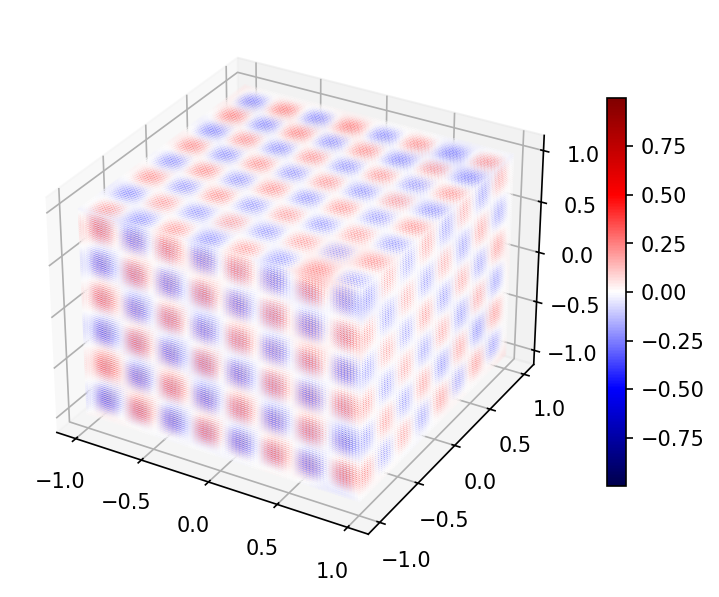} &
		\includegraphics[width=\linewidth]{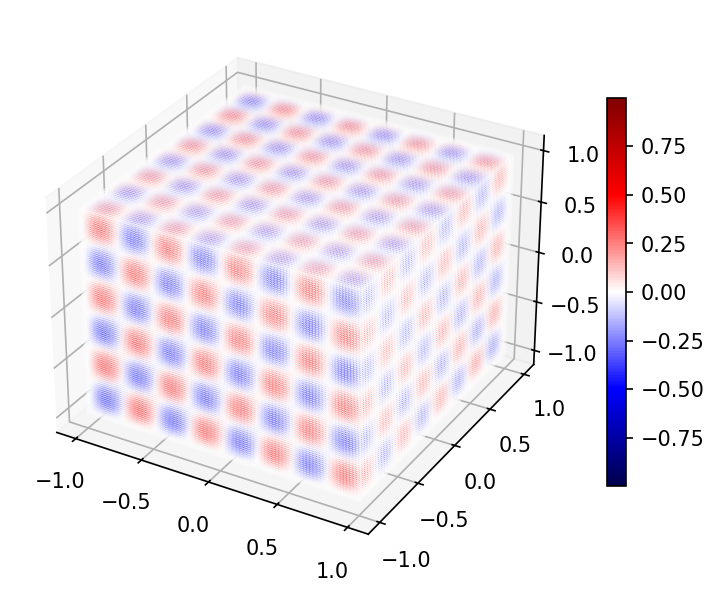} \\
		\rotatebox{90}{Error} &
		\includegraphics[width=\linewidth]{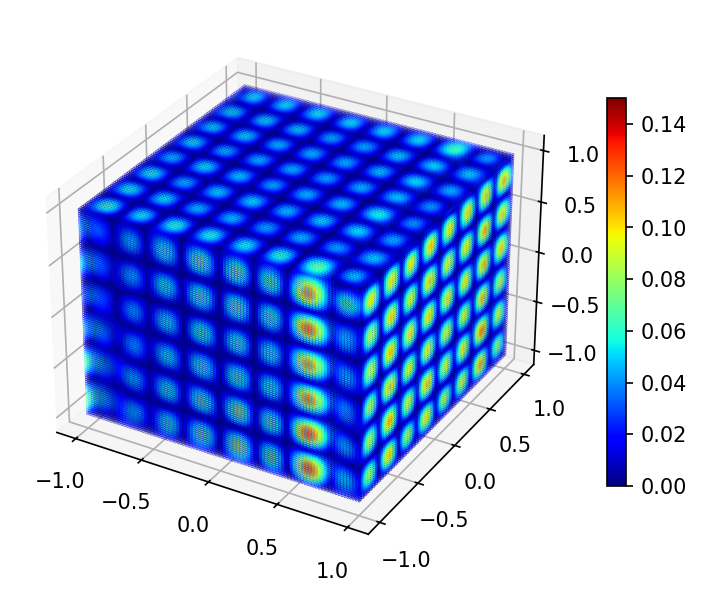} &
		\includegraphics[width=\linewidth]{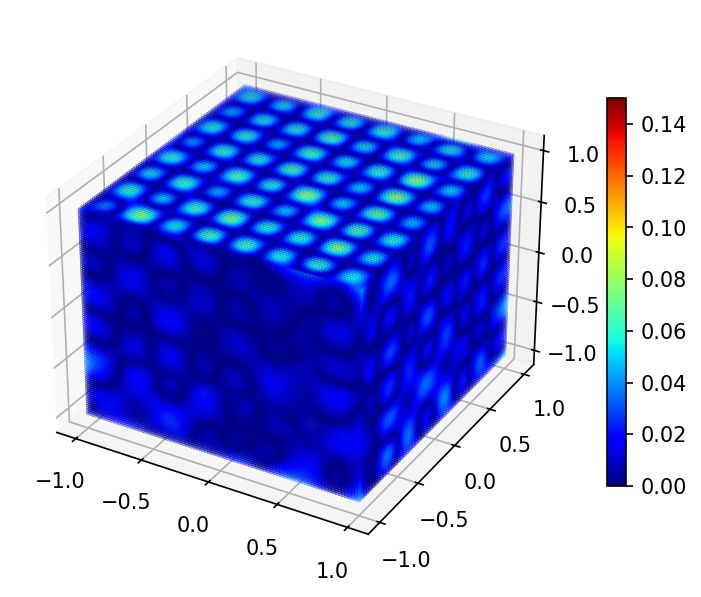} &
		\includegraphics[width=\linewidth]{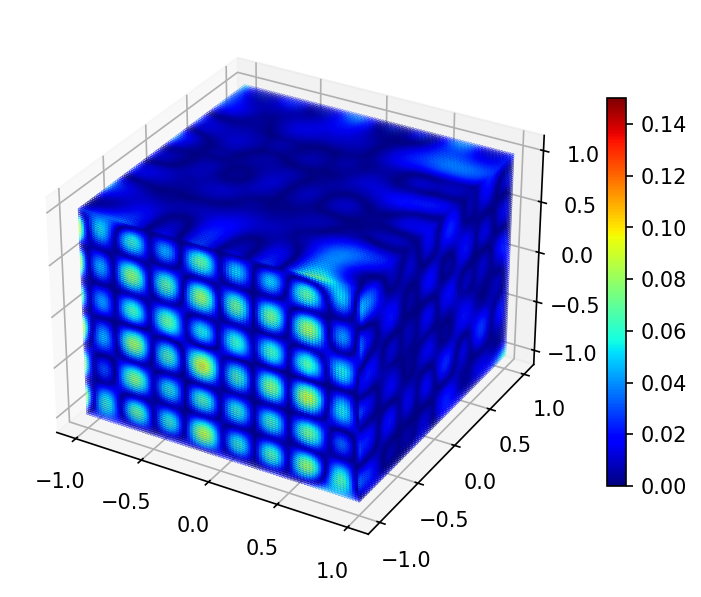} &
		\includegraphics[width=\linewidth]{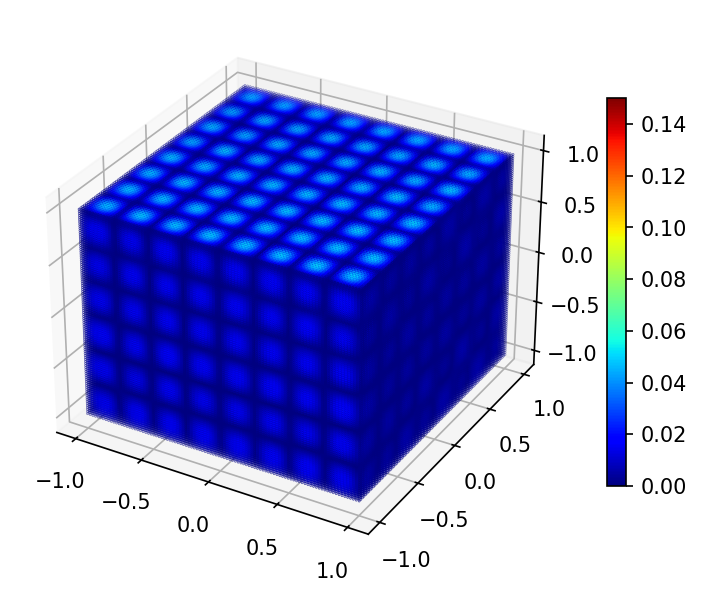} \\
	\end{tabular}
	\caption{Visual comparison of predicted solutions and absolute errors for 3D Helmholtz equation.} \label{fig:helmholtz}
\end{figure}

Table \ref{tab:helmholtz} quantitatively summarizes the numerical performance of SPINN, TT-PINN, Tucker-PINN and HRE-PINN on the 3D Helmholtz equation, including RMSE, relative $L^2$ error and relative $L^\infty$ error. Among all competing tensor-based PINNs, HRE-PINN achieves the lowest approximation errors across all three error metrics: the RMSE reaches $0.005\,021$, the $L^2$ error is reduced to $0.014\,417$, and the $L^\infty$ error is only $0.054\,701$, which is roughly $40\%$ lower than the second-best TT-PINN.
Figure \ref{fig:helmholtz} provides visual comparisons of the predicted solution and absolute error for each method over the cubic domain $\varOmega = [-1,1]^3$. All three baseline tensor-based PINNs exhibit widespread, structured error patterns aligned with the high-frequency sinusoidal components of the manufactured solution, with prominent error hotspots uniformly distributed across the entire computational cube. In contrast, the error of HRE-PINN maintains consistently low magnitude throughout the domain, with nearly no visible high-error regions.
This visual evidence corroborates the quantitative advantage in Table \ref{tab:helmholtz}, verifying that the hierarchical together with rank-evolving mechanism can precisely capture the complex coupling structure of high-frequency elliptic PDE solutions. The consistent superiority in both quantitative metrics and qualitative visualization confirms that the HRE representation alleviates the bottleneck of tensor decompositions with fixed underlying structures for high-frequency wave propagation PDEs.

\subsection{5D Poisson equation}

Poisson equations are the most fundamental elliptic PDEs, describing steady-state phenomena such as electrostatic potential, gravitational fields, and heat conduction. High-dimensional Poisson equations are particularly challenging for traditional numerical methods and PINNs due to the curse of dimensionality, which causes exponential growth in computational cost and memory usage with increasing dimension. We consider the 5D Poisson equation with homogeneous Dirichlet boundary conditions:
\[ \begin{cases}
	- \Delta u(\mat{x}) = f(\mat{x}), & \mat{x} \in \varOmega, \\
	u(\mat{x}) = 0, & \mat{x} \in \partial\varOmega,
\end{cases} \]
where $\varOmega = [-1,1]^5$. We use a simple manufactured analytical solution:
\[ u(\mat{x}) = \sum_{i=1}^{5} \sin\left(\frac{\pi}{2} x_i\right), \qquad \mat{x} = (x_1, x_2, x_3, x_4, x_5), \]
with the corresponding source term
\[ f(\mat{x}) = \frac{\pi^2}{4} \sum_{i=1}^{5} \sin\left(\frac{\pi}{2} x_i\right). \]
This benchmark specifically evaluates the high-dimensional scalability of HRE-PINN architecture.

\begin{table}[width=.7\linewidth,cols=3,pos=!htb]
	\caption{Performance comparison of different methods for 5D Poisson equation} \label{tab:poisson}
	\begin{tabular*}{\tblwidth}{@{}LLLL@{}}
		\toprule
		Method & RMSE & $L^2$ error & $L^\infty$ error \\
		\midrule
		SPINN & 0.001\,899 & 0.000\,586 & 0.003\,586 \\
		TT-PINN & 0.001\,997 & 0.000\,616 & 0.004\,052 \\
		Tucker-PINN & 0.002\,125 & 0.000\,656 & 0.003\,329 \\
		HRE-PINN & 0.001\,078 & 0.000\,333 & 0.001\,675 \\
		\bottomrule
	\end{tabular*}
\end{table}

\begin{figure}[pos=!htb]
	\centering
	\begin{tabular}{@{} @{\extracolsep{\fill}} m{1em} *{4}{>{\centering\arraybackslash}m{0.2\linewidth}} @{}}
		& SPINN & TT-PINN & Tucker-PINN & HRE-PINN \\
		\rotatebox{90}{Predicted} &
		\includegraphics[width=\linewidth]{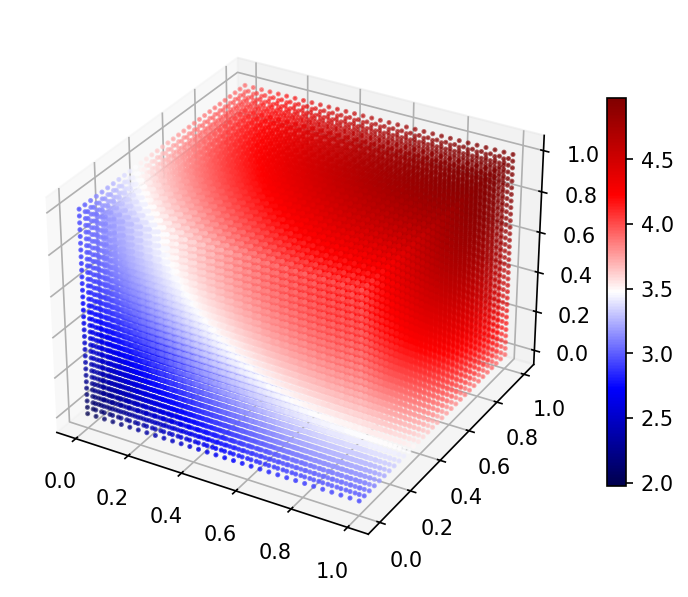} &
		\includegraphics[width=\linewidth]{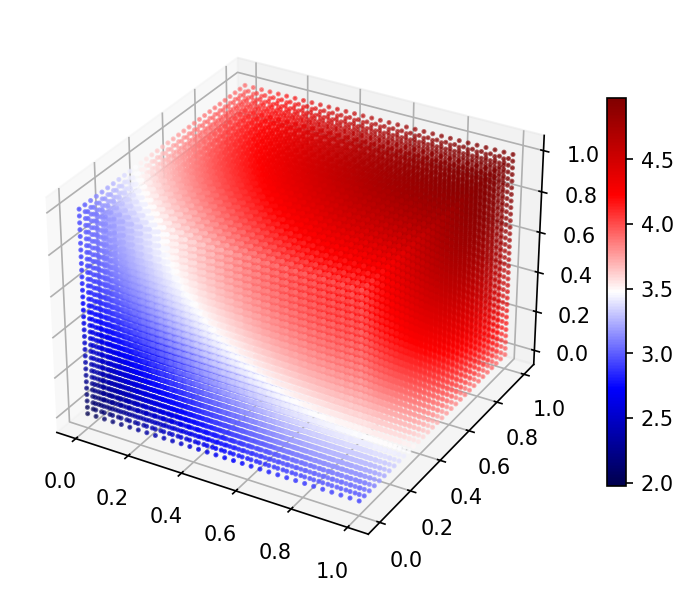} &
		\includegraphics[width=\linewidth]{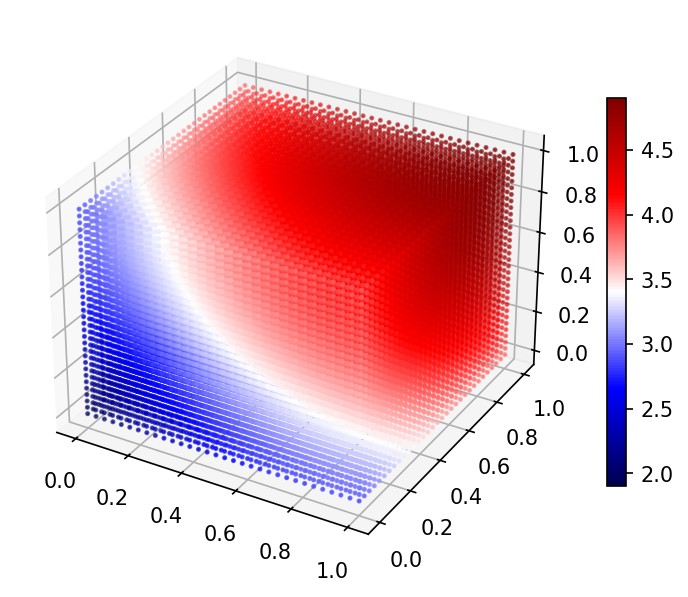} &
		\includegraphics[width=\linewidth]{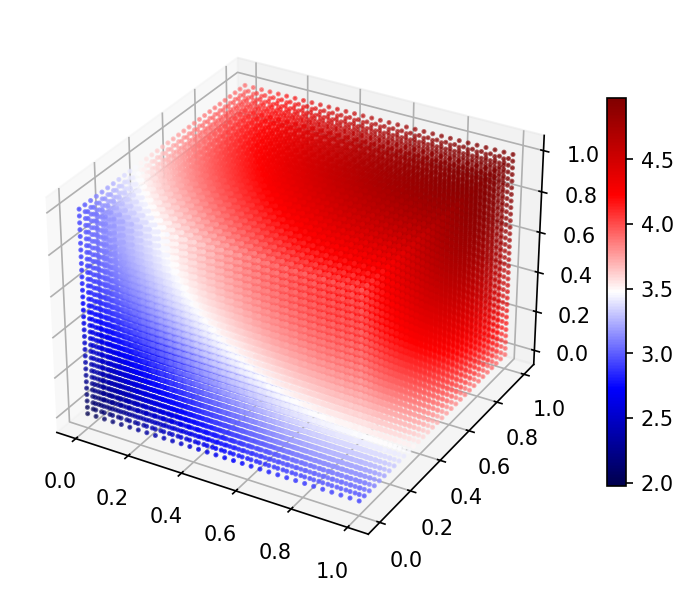} \\
		\rotatebox{90}{Error} &
		\includegraphics[width=\linewidth]{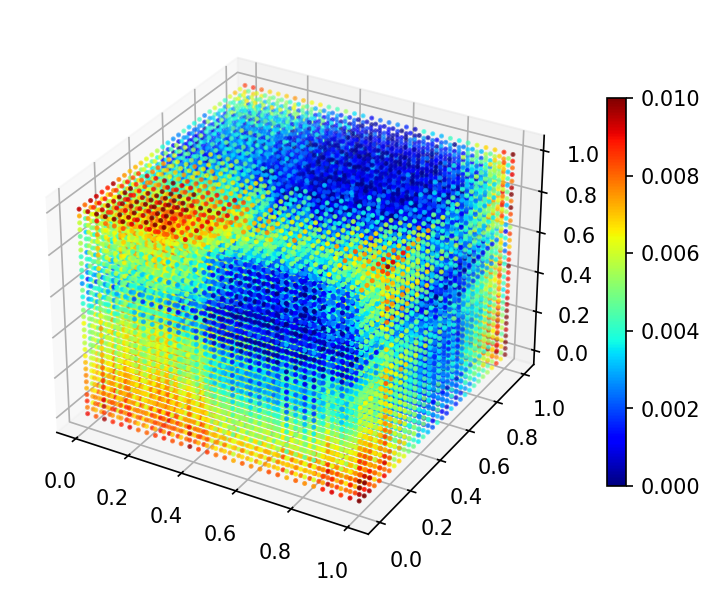} &
		\includegraphics[width=\linewidth]{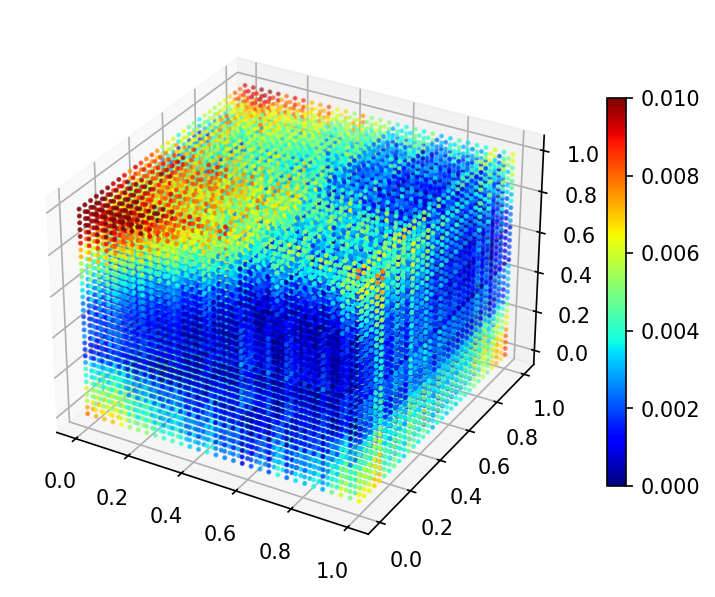} &
		\includegraphics[width=\linewidth]{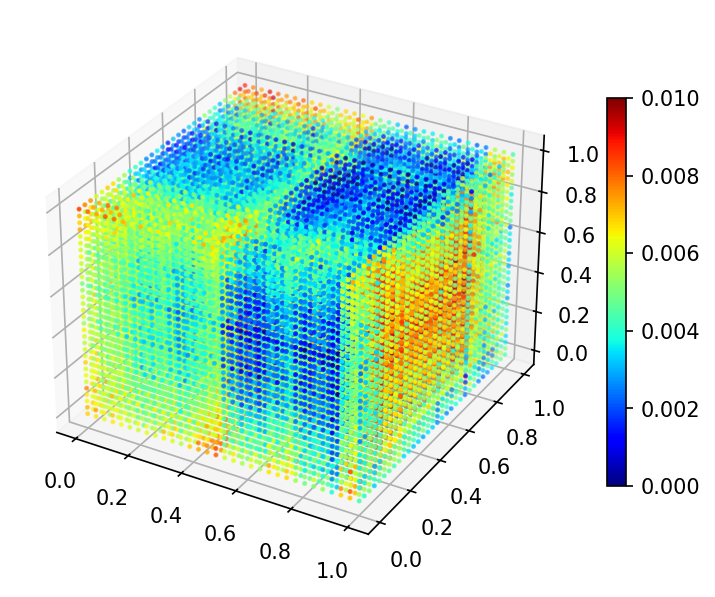} &
		\includegraphics[width=\linewidth]{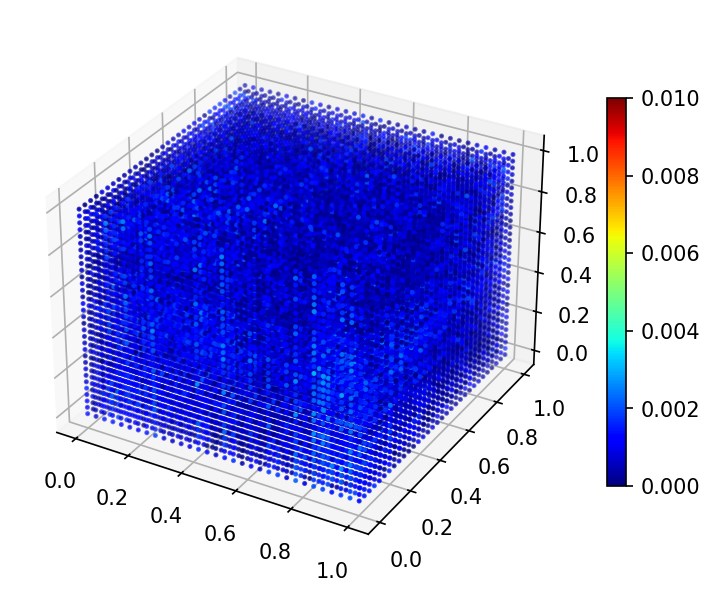} \\
	\end{tabular}
	\caption{Visual comparison of predicted solutions and absolute errors for 5D Poisson equation, sliced at $x_4 = x_5 = 0.9$.} \label{fig:poisson}
\end{figure}

Table \ref{tab:poisson} reports the quantitative performance of SPINN, TT-PINN, Tucker-PINN and HRE-PINN on the 5D Poisson equation, including RMSE, relative $L^2$ error and relative $L^\infty$ error. This high-dimensional case clearly demonstrates the strong scalability of our hierarchical framework against the curse of dimensionality. Compared with all baseline tensor-based PINNs, HRE-PINN achieves the optimal accuracy on all error metrics: its RMSE is $0.001\,078$, the $L^2$ error drops to $0.000\,333$, and the $L^\infty$ error equals $0.001\,675$, which outperforms the second-best SPINN by approximately $50\%$.
Figure \ref{fig:poisson} visualizes the predicted solution and absolute error for each method. For SPINN, TT-PINN and Tucker-PINN, the error fields present obvious widespread high-value regions across the entire hypercube domain, indicating insufficient expressive power of fixed-topology tensor decompositions to capture the multi-dimensional coupling relations in 5D elliptic solutions. By contrast, the error magnitude of HRE-PINN is suppressed to a lower level.
This qualitative observation aligns perfectly with the quantitative data in Table \ref{tab:poisson}, validating that the hierarchical structure of HRE can automatically determining the suitable ranks and flexible underlying structure without manual hyperparameter tuning. For high-dimensional PDEs where traditional discretization and standard PINNs are hindered by exponential computational cost, the rank-evolving mechanism effectively eliminates redundant model degrees of freedom, enabling precise approximation of high-dimensional physical fields.

\subsection{(2+1)D Klein-Gordon equation}

Klein-Gordon equations are nonlinear hyperbolic PDEs that arise in relativistic quantum mechanics, particle physics, and nonlinear optics. They describe the propagation of relativistic waves and exhibits rich dynamical behavior including wave dispersion and nonlinear interactions. The (2+1)D formulation presents significant challenges due to the combination of second-order temporal derivatives, second-order spatial derivatives, and nonlinear terms. We solve the inhomogeneous Klein-Gordon equation:
\[ \begin{cases}
	\partial_{tt} u(x,y,t) - \Delta u(x,y,t) + u^2(x,y,t) = f(x,y,t), & (x,y) \in \varOmega,\ t \in \varGamma, \\
	u(x,y,t) = b(x,y,t), & (x,y) \in \partial\varOmega,\ t \in \varOmega, \\
	u(x,y,0) = x + y, & (x,y) \in \varOmega,
\end{cases} \]
where $\varOmega = [-1,1]^2$, $\varGamma = [0,10]$. The exact analytical solution is given by:
\[ u(x,y,t) = (x + y) \cos 2t + xy \sin 2t, \]
from which the source term $f(x,y,t)$ and boundary condition $b(x,y,t)$ are derived. This benchmark assesses performance of HRE-PINN on time-dependent, second-order hyperbolic systems with nonlinearities.

\begin{table}[width=.7\linewidth,cols=4,pos=!htb]
	\caption{Performance comparison of different methods for (2+1)D Klein-Gordon equation} \label{tab:klein-gordon}
	\begin{tabular*}{\tblwidth}{@{}LLLL@{}}
		\toprule
		Method & RMSE & $L^2$ error & $L^\infty$ error \\
		\midrule
		SPINN & 0.002\,440 & 0.003\,841 & 0.008\,030 \\
		TT-PINN & 0.007\,433 & 0.011\,702 & 0.040\,222 \\
		Tucker-PINN & 0.008\,823 & 0.013\,890 & 0.096\,783 \\
		HRE-PINN & 0.000\,468 & 0.000\,737 & 0.001\,949 \\
		\bottomrule
	\end{tabular*}
\end{table}

\begin{figure}[pos=!htb]
	\centering
	\begin{tabular}{@{} @{\extracolsep{\fill}} m{1em} *{4}{>{\centering\arraybackslash}m{0.2\linewidth}} @{}}
		& SPINN & TT-PINN & Tucker-PINN & HRE-PINN \\
		\rotatebox{90}{Predicted} &
		\includegraphics[width=\linewidth]{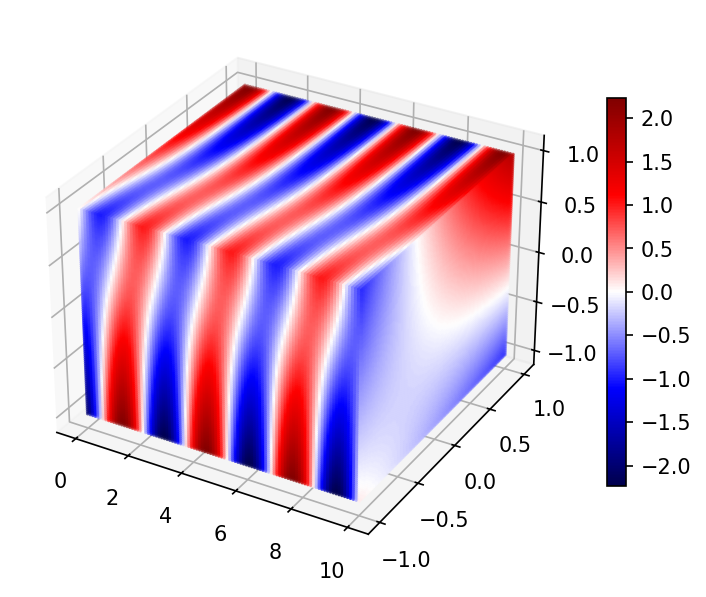} &
		\includegraphics[width=\linewidth]{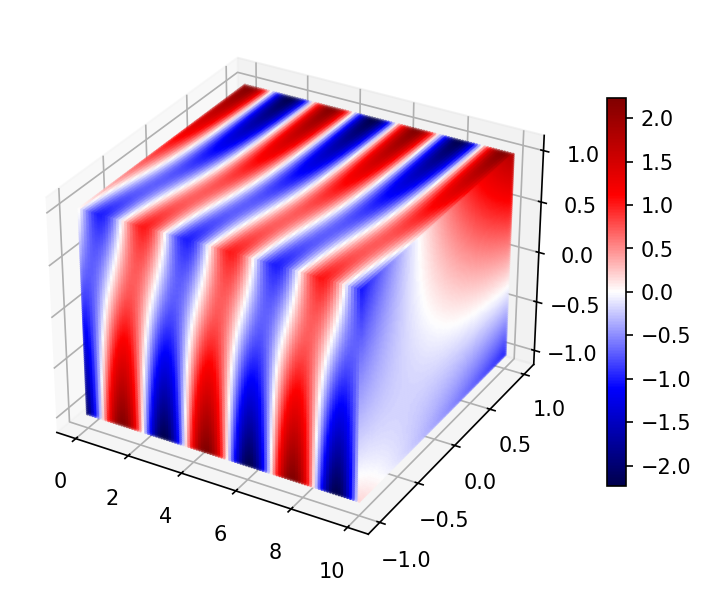} &
		\includegraphics[width=\linewidth]{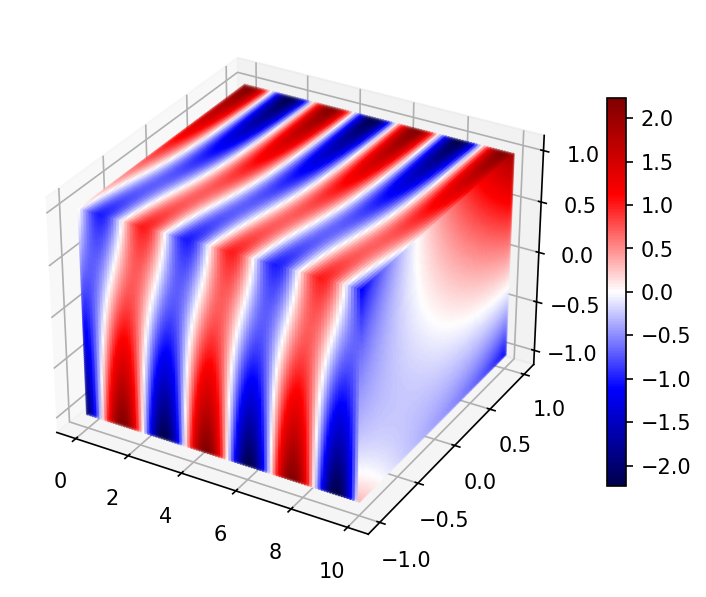} &
		\includegraphics[width=\linewidth]{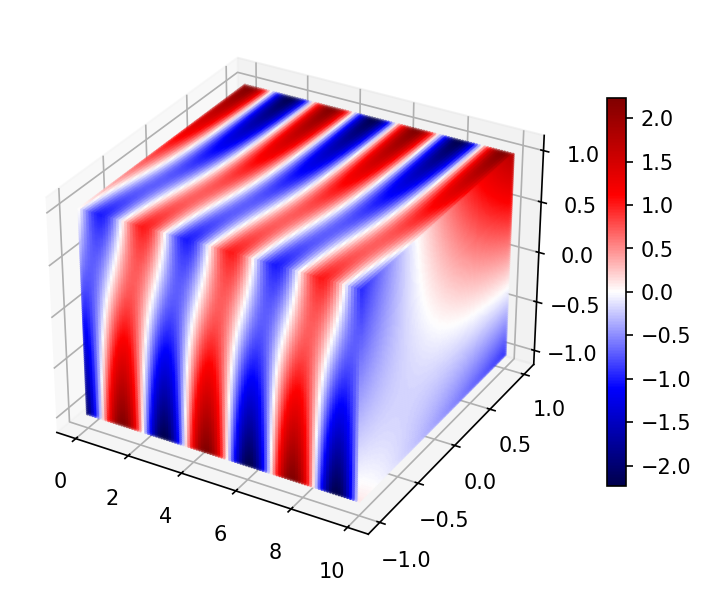} \\
		\rotatebox{90}{Error} &
		\includegraphics[width=\linewidth]{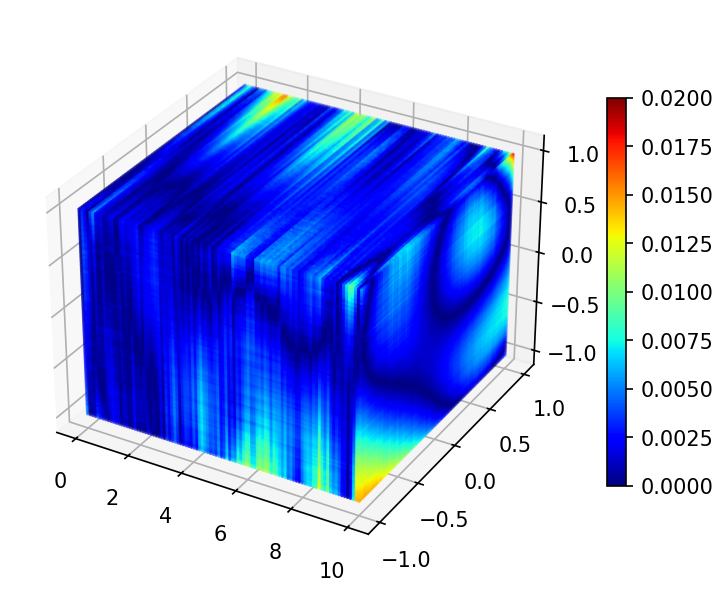} &
		\includegraphics[width=\linewidth]{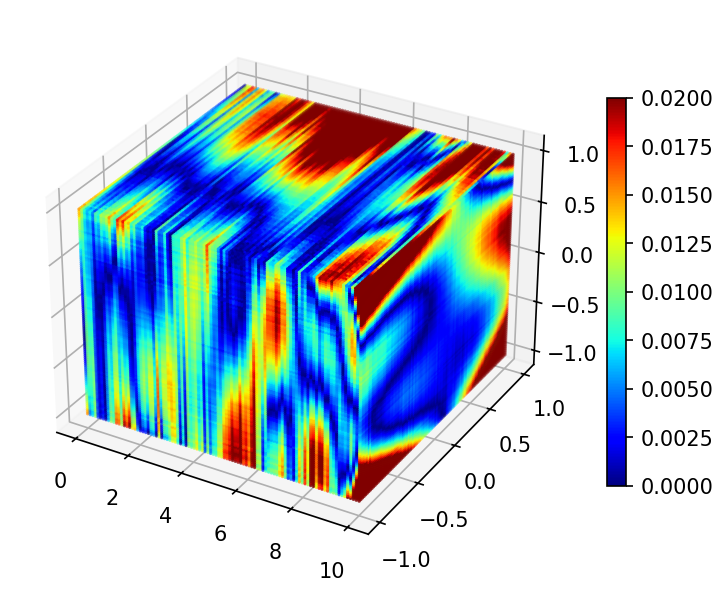} &
		\includegraphics[width=\linewidth]{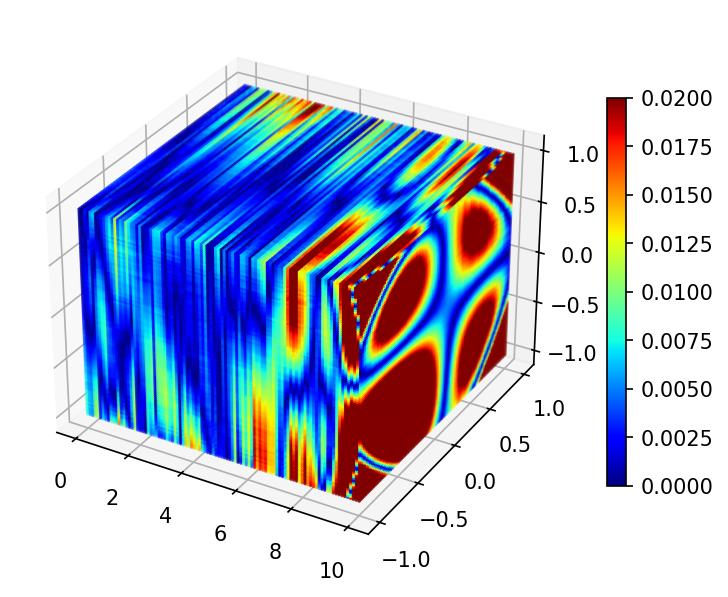} &
		\includegraphics[width=\linewidth]{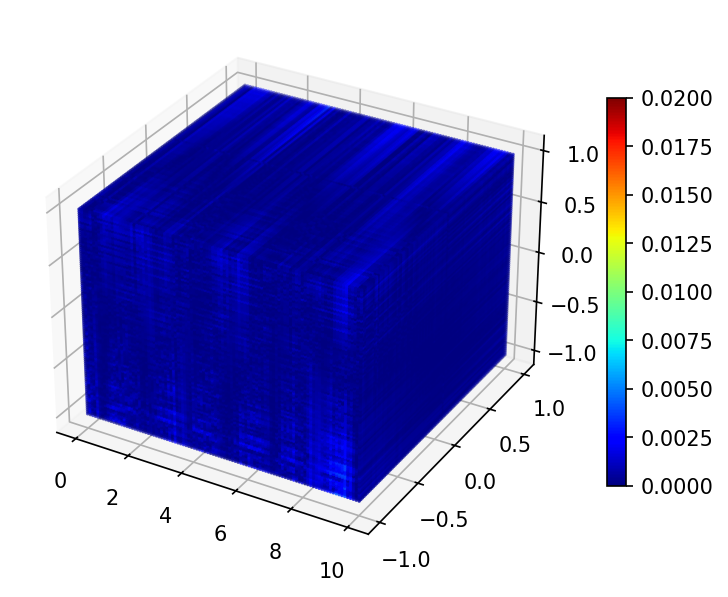} \\
	\end{tabular}
	\caption{Visual comparison of predicted solutions and absolute errors for (2+1)D Klein-Gordon equation.} \label{fig:klein-gordon}
\end{figure}

Table \ref{tab:klein-gordon} quantitatively compares the numerical performance of SPINN, TT-PINN, Tucker-PINN and HRE-PINN on the nonlinear time-dependent (2+1)D Klein-Gordon equation, listing the number of RMSE, relative $L^2$ error, and relative $L_\infty$ error. This hyperbolic PDE with second-order temporal derivatives and quadratic nonlinear terms poses a severe challenge to conventional tensor-based PINNs, yet HRE-PINN achieves the best accuracy across all error metrics. Specifically, the RMSE of HRE-PINN is merely $0.000\,468$, with a relative $L^2$ error of $0.000\,737$ and a relative $L^\infty$ error of $0.001\,949$, which reduces the best baseline (SPINN) by roughly $80\%$.
Figure \ref{fig:klein-gordon} visualizes predicted solution and absolute error over the spatio-temporal domain for all comparative methods. SPINN, TT-PINN and Tucker-PINN all exhibit prominent errors distributed throughout the entire space-time cube. In stark contrast, the error of HRE-PINN maintains an extremely low amplitude across the whole computational domain, with almost no distinct high-error regions visible.
This qualitative observation is fully consistent with the quantitative results in Table \ref{tab:klein-gordon}, and verifies that the hierarchical design and flexible underlying structure are effective in characterizing complex space-time correlations embedded in nonlinear hyperbolic PDE solutions. The rank-evolving mechanism avoids manual rank tuning and effectively mitigates training instability, demonstrating strong applicability to time-dependent relativistic wave problems with intricate dynamical behaviors.

\subsection{(2+1)D flow mixing equation}

The flow mixing problem models the mixing of two fluids at their interface and is widely used in chemical engineering and fluid dynamics. The problem is challenging due to its spatially varying velocity coefficients and sharp interface structures, which require high resolution to capture accurately. We consider the following formulation:
\[ \begin{cases}
	\partial_t u(x,y,t) + a(x,y) \partial_x u(x,y,t) + b(x,y) \partial_y u(x,y,t) = 0, & (x,y) \in \varOmega,\ t \in \varGamma, \\
	u(x,y,0) = - \tanh \frac{y}{2}, & (x,y) \in \varOmega, \\
	u(x,y,t) = b(x,y,t), & (x,y) \in \partial\varOmega,\ t \in \varGamma,
\end{cases} \]
where $\varOmega = [-4,4]^2$, $\varGamma = [0,4]$, and the spatially varying velocity coefficients are defined as:
\[ a(x,y) = -\frac{v_t}{v_{t,\max}} \frac{y}{r}, \qquad b(x,y) = \frac{v_t}{v_{t,\max}} \frac{x}{r}, \qquad v_t = \frac{\tanh r}{\cosh^2 r}, \qquad r = \sqrt{x^2 + y^2}, \qquad v_{t,\max} = 0.385. \]
The exact analytical solution for this problem is
\[ u(x,y,t) = - \tanh\left( \frac{y}{2} \cos\left( \frac{1}{r} \frac{v_t}{v_{t,\max}} t\right) - \frac{x}{2} \sin\left( \frac{1}{r} \frac{v_t}{v_{t,\max}} t\right) \right), \]
and the Dirichlet boundary condition $b(x,y,t)$ is extracted from this solution.
This benchmark tests the ability of HRE-PINN to handle sharp interface dynamics.

\begin{table}[width=.7\linewidth,cols=3,pos=!htb]
	\caption{Performance comparison of different methods for (2+1)D flow mixing equation} \label{tab:flow mixing}
	\begin{tabular*}{\tblwidth}{@{}LLLL@{}}
		\toprule
		Method & RMSE & $L^2$ error & $L^\infty$ error \\
		\midrule
		SPINN & 0.024\,408 & 0.033\,809 & 0.279\,922 \\
		TT-PINN & 0.018\,491 & 0.025\,613 & 0.220\,849 \\
		Tucker-PINN & 0.024\,644 & 0.034\,136 & 0.272\,519 \\
		HRE-PINN & 0.002\,487 & 0.003\,445 & 0.032\,788 \\
		\bottomrule
	\end{tabular*}
\end{table}

\begin{figure}[pos=!htb]
	\centering
	\begin{tabular}{@{} @{\extracolsep{\fill}} m{1em} *{4}{>{\centering\arraybackslash}m{0.2\linewidth}} @{}}
		& SPINN & TT-PINN & Tucker-PINN & HRE-PINN \\
		\rotatebox{90}{Predicted} &
		\includegraphics[width=\linewidth]{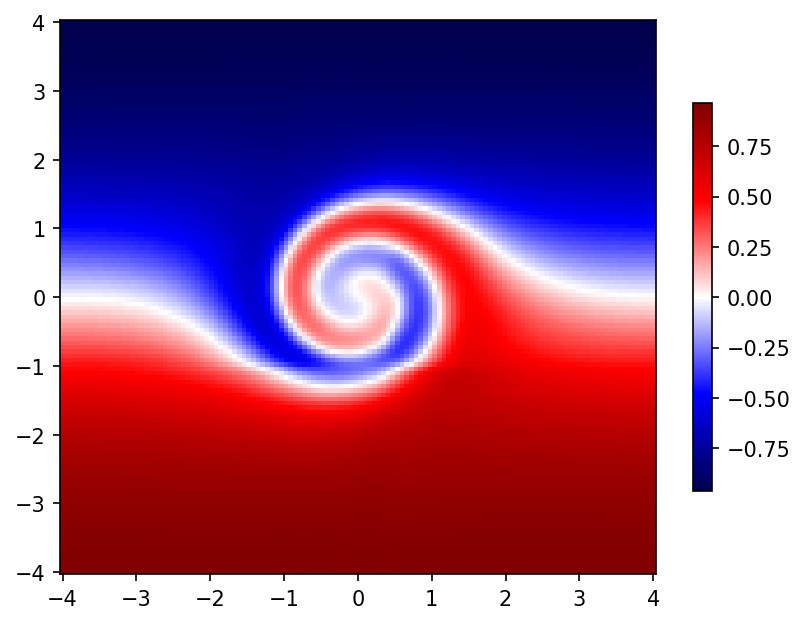} &
		\includegraphics[width=\linewidth]{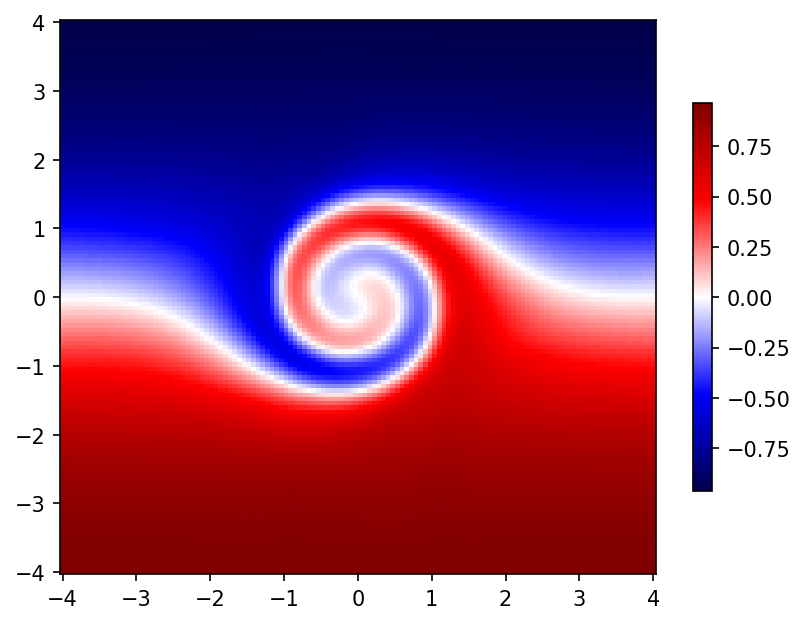} &
		\includegraphics[width=\linewidth]{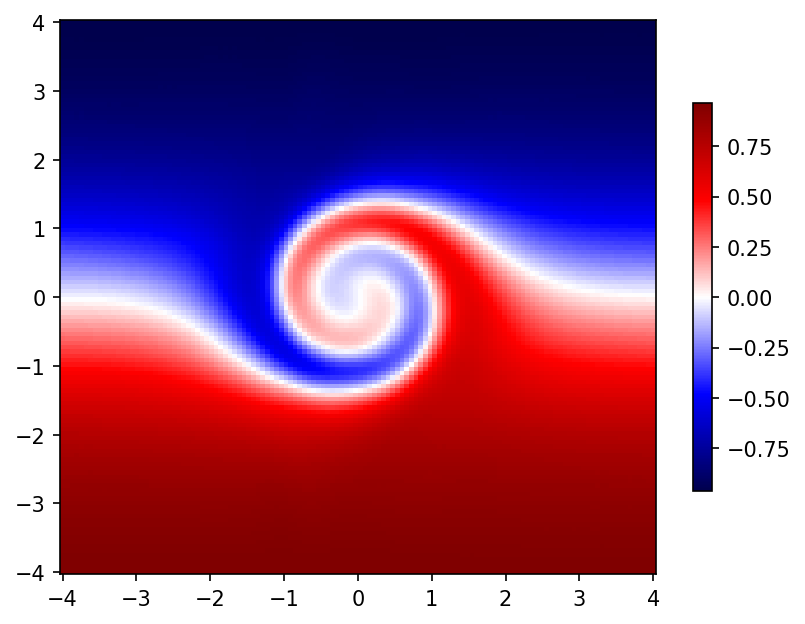} &
		\includegraphics[width=\linewidth]{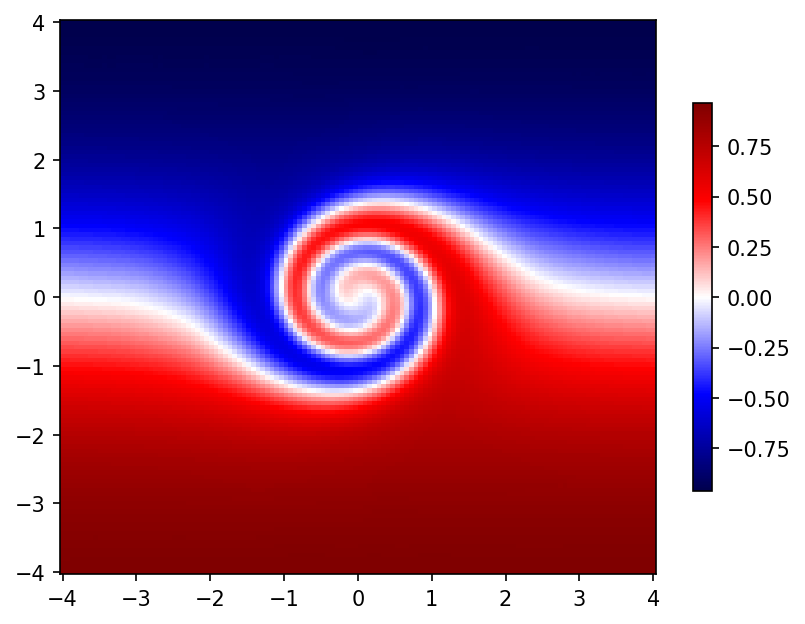} \\
		\rotatebox{90}{Error} &
		\includegraphics[width=\linewidth]{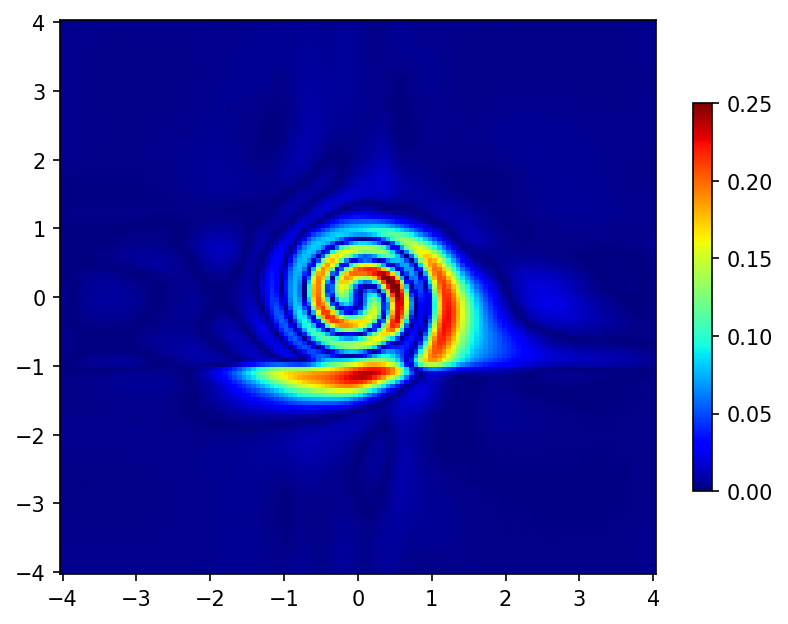} &
		\includegraphics[width=\linewidth]{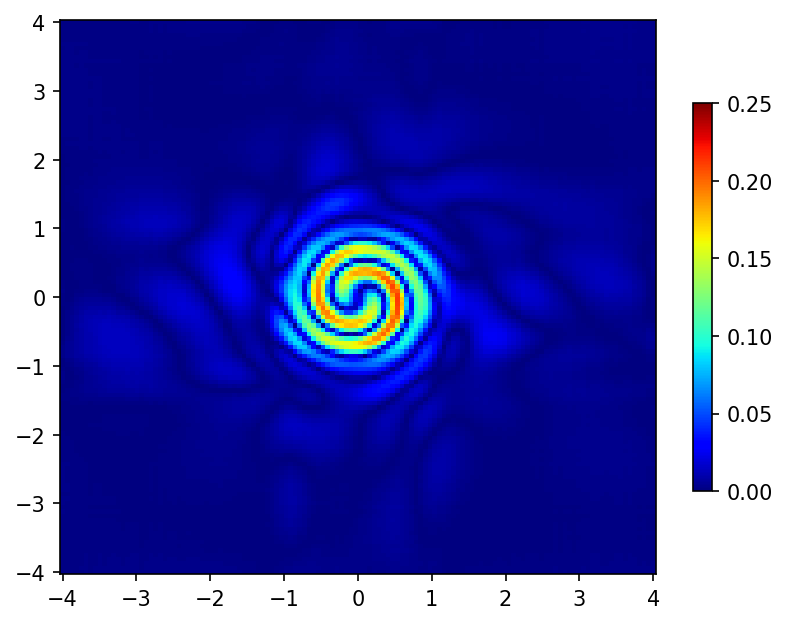} &
		\includegraphics[width=\linewidth]{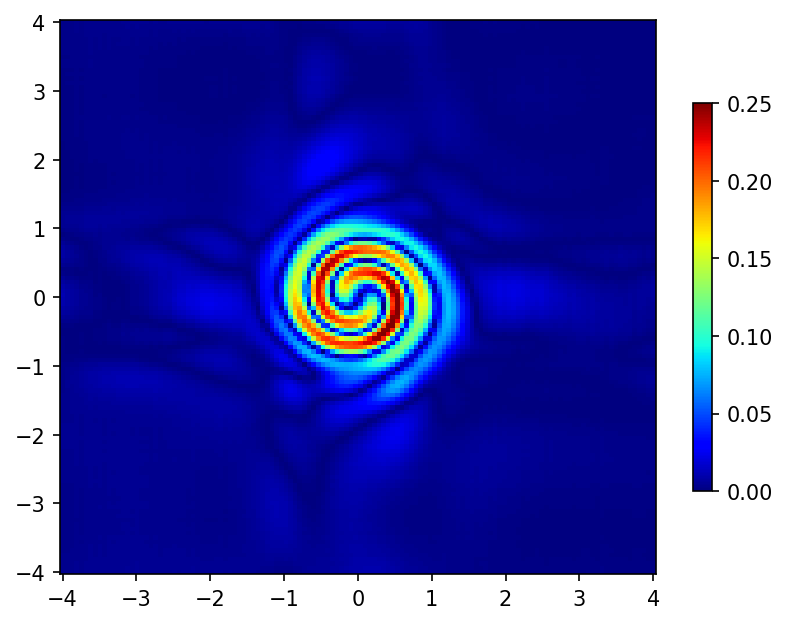} &
		\includegraphics[width=\linewidth]{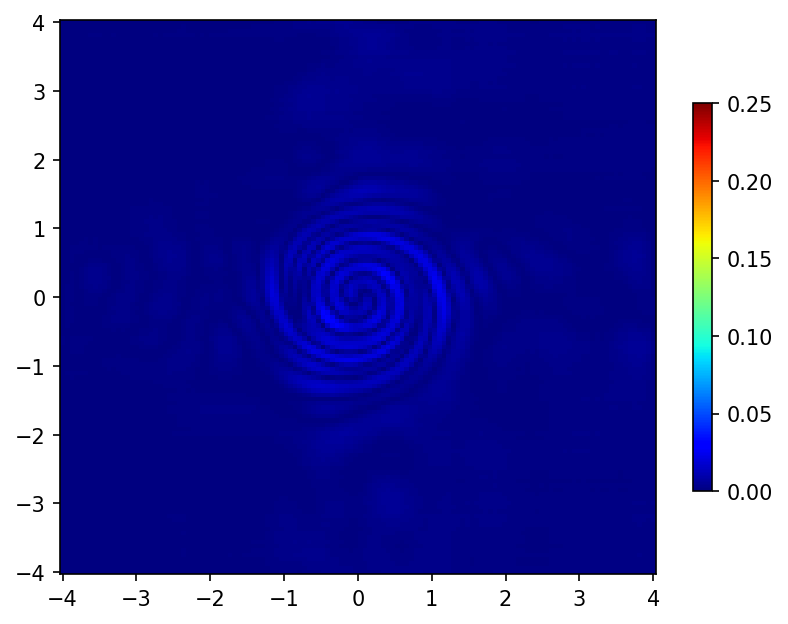} \\
	\end{tabular}
	\caption{Visual comparison of predicted solutions and absolute errors for (2+1)D flow mixing equation at $t=4$.} \label{fig:flow mixing}
\end{figure}

Table \ref{tab:flow mixing} presents the quantitative performance metrics of SPINN, TT-PINN, Tucker-PINN and HRE-PINN on the advection-dominated flow mixing benchmark, including RMSE, relative $L^2$ error, and relative $L^\infty$ error. This PDE with spatially variable rotational velocity fields and sharp fluid interface structures severely tests the representation capacity of tensor-based PINNs, and HRE-PINN achieves overwhelming accuracy advantages over all baseline methods. Numerically, HRE-PINN yields an RMSE of $0.002\,487$, a relative $L^2$ error of $0.003\,445$ and a relative $L^\infty$ error of $0.032\,788$, reducing the error of the second-best TT-PINN by approximately $80\%$.
Figure \ref{fig:flow mixing} provides intuitive visual comparisons of the predicted solution and absolute error for each competing method across the spatial domain at $t=4$. SPINN, TT-PINN and Tucker-PINN all generate prominent high-error bands concentrated along the twisted mixing interface; such concentrated error artifacts reveal that tensor decompositions with fixed underlying structures lack sufficient flexibility to capture complex correlations. In contrast, the error distribution of HRE-PINN remains uniformly close to zero throughout the entire domain.
This qualitative observation strongly supports the quantitative results in Table \ref{tab:flow mixing}, and validates that the hierarchical framework equipped with rank-evolving mechanism can automatically revealing the suitable underlying structure to resolve steep flow interfaces without predefining structures manually. For advection-dominated fluid transport problems featuring sharp transition layers, the customized tensor network decomposition effectively characterizes complex cross-dimensional coupling, while the sparsity-driven rank-evolving mechanism prevents over-parameterization and delivers high-fidelity approximations of mixing dynamics.

\subsection{(2+1)D Navier-Stokes equation}

Navier-Stokes equations are governing equations of fluid dynamics, describing the motion of viscous incompressible fluids. The (2+1)D decaying turbulence problem is a challenging benchmark that tests the ability of numerical methods to capture complex, chaotic flow structures and nonlinear interactions between vortices. We solve the vorticity formulation of the equations on a periodic domain:
\[ \begin{cases}
	\partial_t \omega(x,y,t) + \mat{u}(x,y,t) \cdot \nabla \omega(x,y,t) = \nu \Delta \omega(x,y,t), & (x,y) \in \varOmega,\ t \in \varGamma, \\
	\nabla \cdot \mat{u}(x,y,t) = 0, & (x,y) \in \varOmega,\ t \in \varGamma, \\
	\omega(x,y,0) = \omega_0(x,y), & (x,y) \in \varOmega,
\end{cases} \]
where $\varOmega = [0,2\pi]^2$, $\varGamma = [0, 1]$ and $\mat{u}(x,y,t) \in \mathbb{R}^2$ is the velocity field, $\omega = \nabla \times \mat{u}$ is the vorticity, and $\nu = 0.01$ is the kinematic viscosity. The initial vorticity field $\omega_0(\mathbf{x})$ is generated as a Gaussian random field with a maximum velocity magnitude of 5, following the setup in \cite{wang2024respecting}. We generate a high-resolution reference solution using the JAX-CFD solver \cite{kochkov2021machine} with a resolution of $128 \times 128 \times 100$. This benchmark represents the most challenging test in our suite, evaluating the ability of HRE-PINN to handle strong nonlinearities and chaotic dynamics.

\begin{table}[width=.7\linewidth,cols=3,pos=!htb]
	\caption{Performance comparison of different methods for vorticity‑field predictions of (2+1)D Navier-Stokes equation} \label{tab:navier-stokes}
	\begin{tabular*}{\tblwidth}{@{}LLLL@{}}
		\toprule
		Method & RMSE & $L^2$ error & $L^\infty$ error \\
		\midrule
		SPINN & 0.762\,091 & 0.134\,820 & 0.207\,020 \\
		TT-PINN & 1.320\,352 & 0.233\,581 & 0.281\,729 \\
		Tucker-PINN & 1.003\,436 & 0.177\,516 & 0.235\,614 \\
		HRE-PINN & 0.414\,919 & 0.073\,403 & 0.159\,532 \\
		\bottomrule
	\end{tabular*}
\end{table}

\begin{figure}[pos=!htb]
	\centering
	\begin{tabular}{@{} @{\extracolsep{\fill}} m{1em} *{4}{>{\centering\arraybackslash}m{0.2\linewidth}} @{}}
		& SPINN & TT-PINN & Tucker-PINN & HRE-PINN \\
		\rotatebox{90}{Predicted} &
		\includegraphics[width=\linewidth]{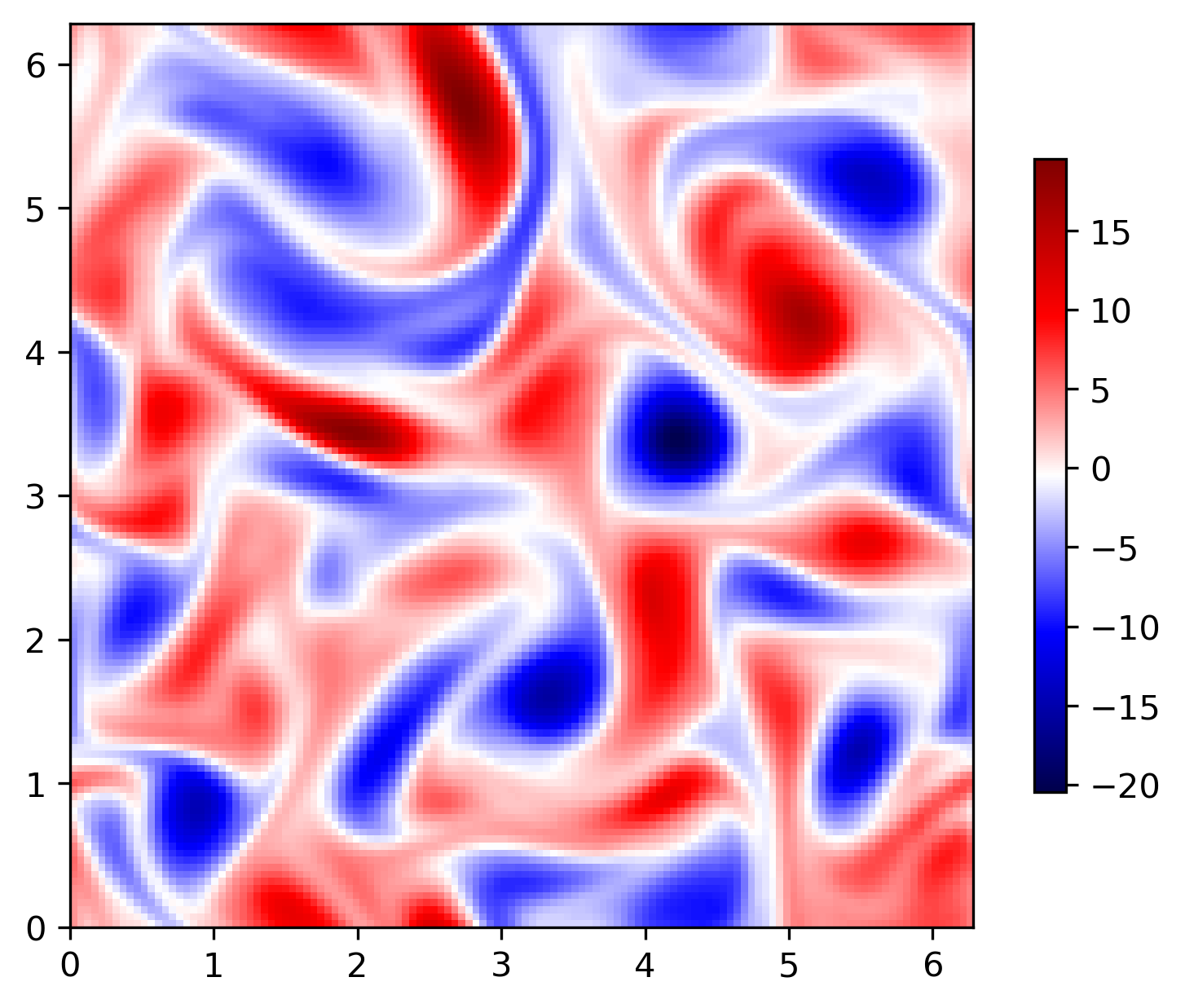} &
		\includegraphics[width=\linewidth]{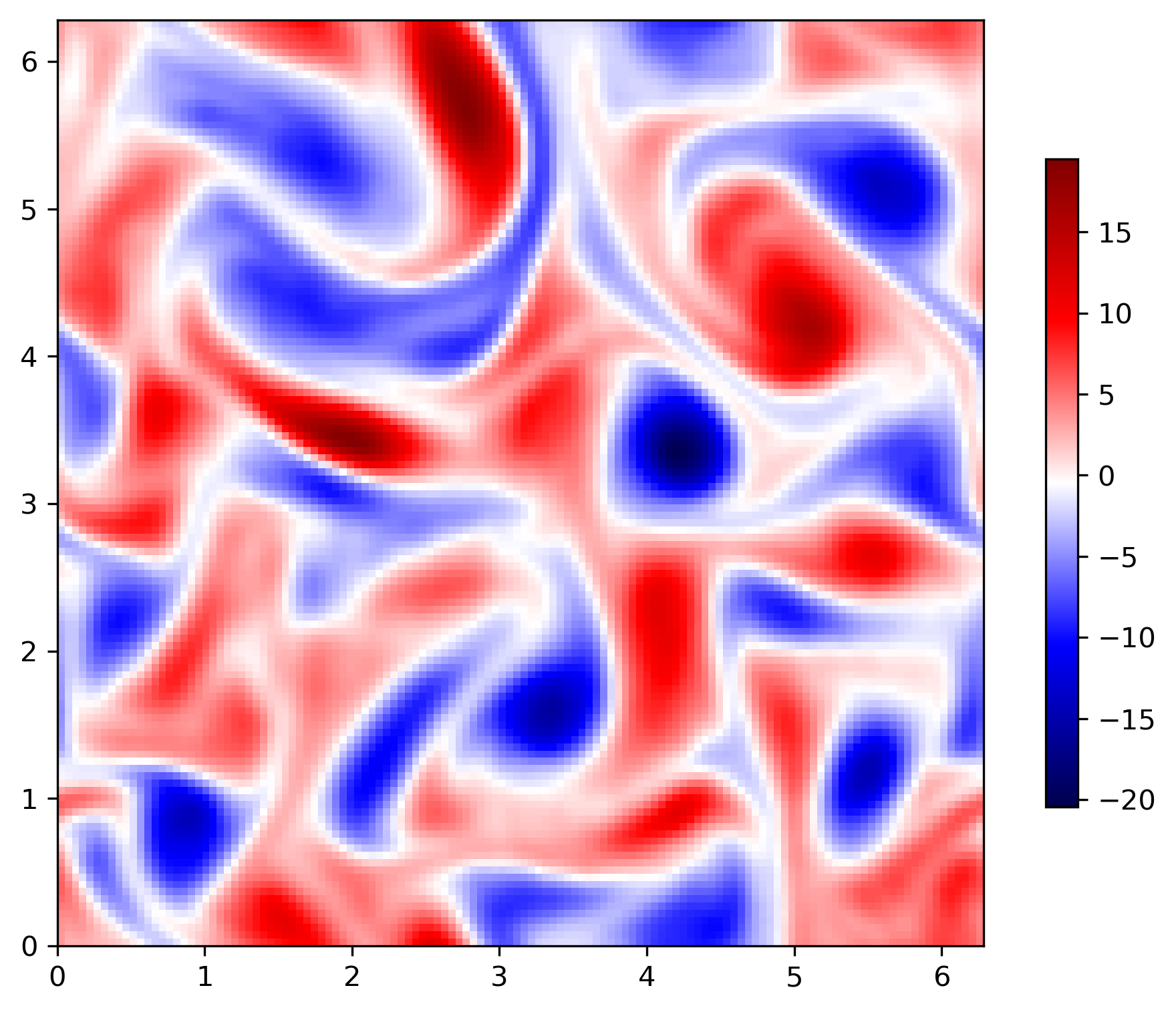} &
		\includegraphics[width=\linewidth]{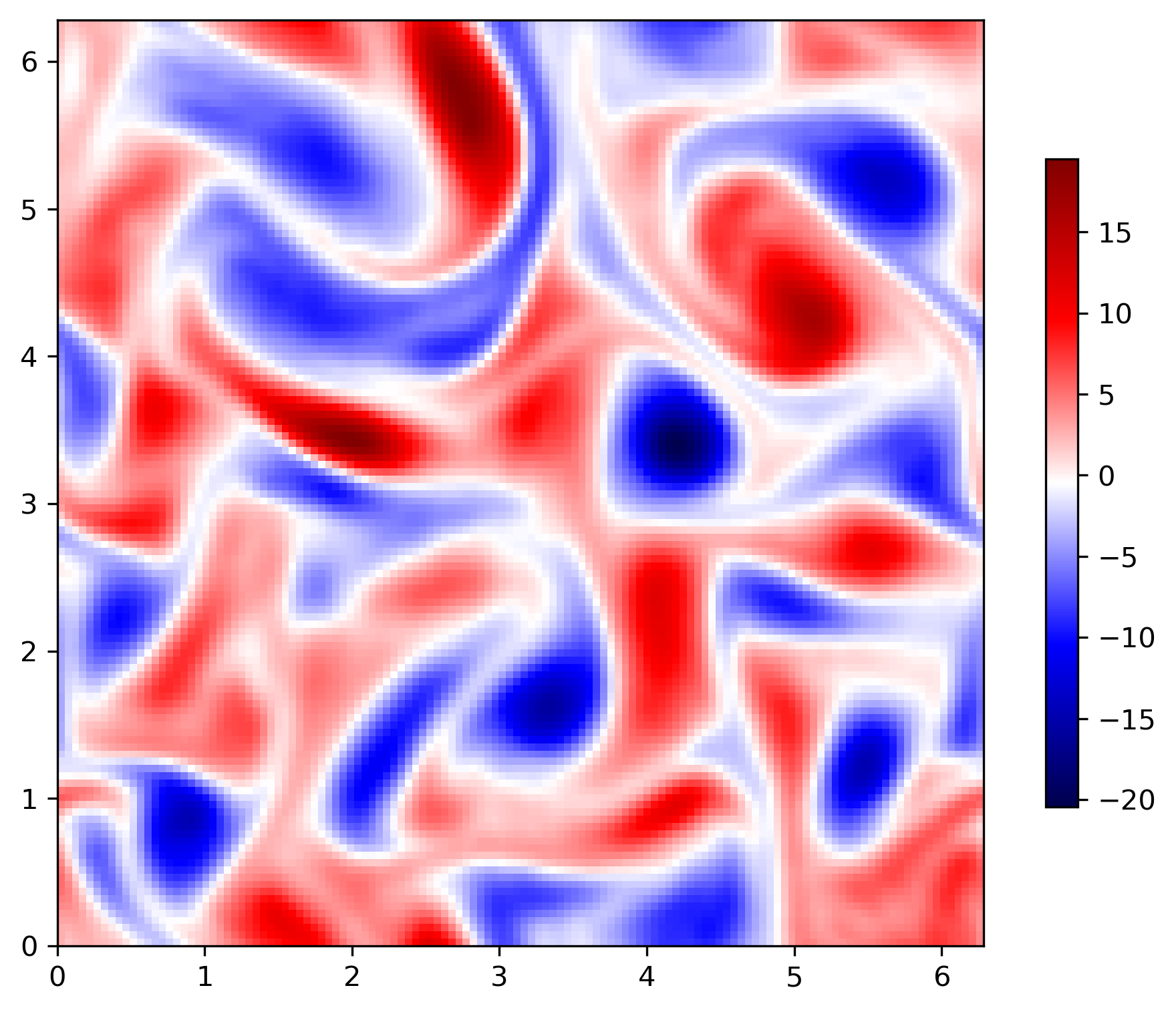} &
		\includegraphics[width=\linewidth]{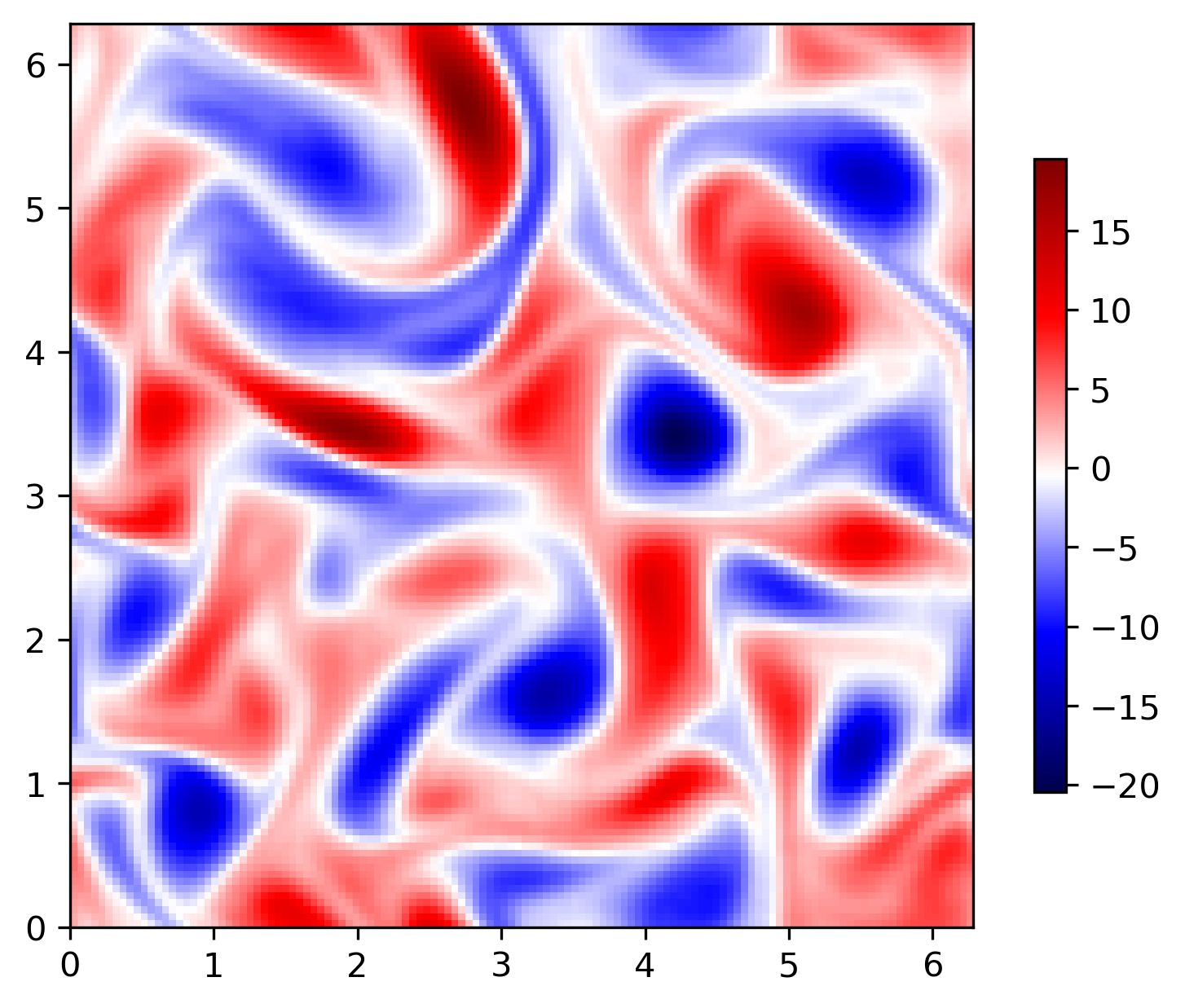} \\
		\rotatebox{90}{Error} &
		\includegraphics[width=\linewidth]{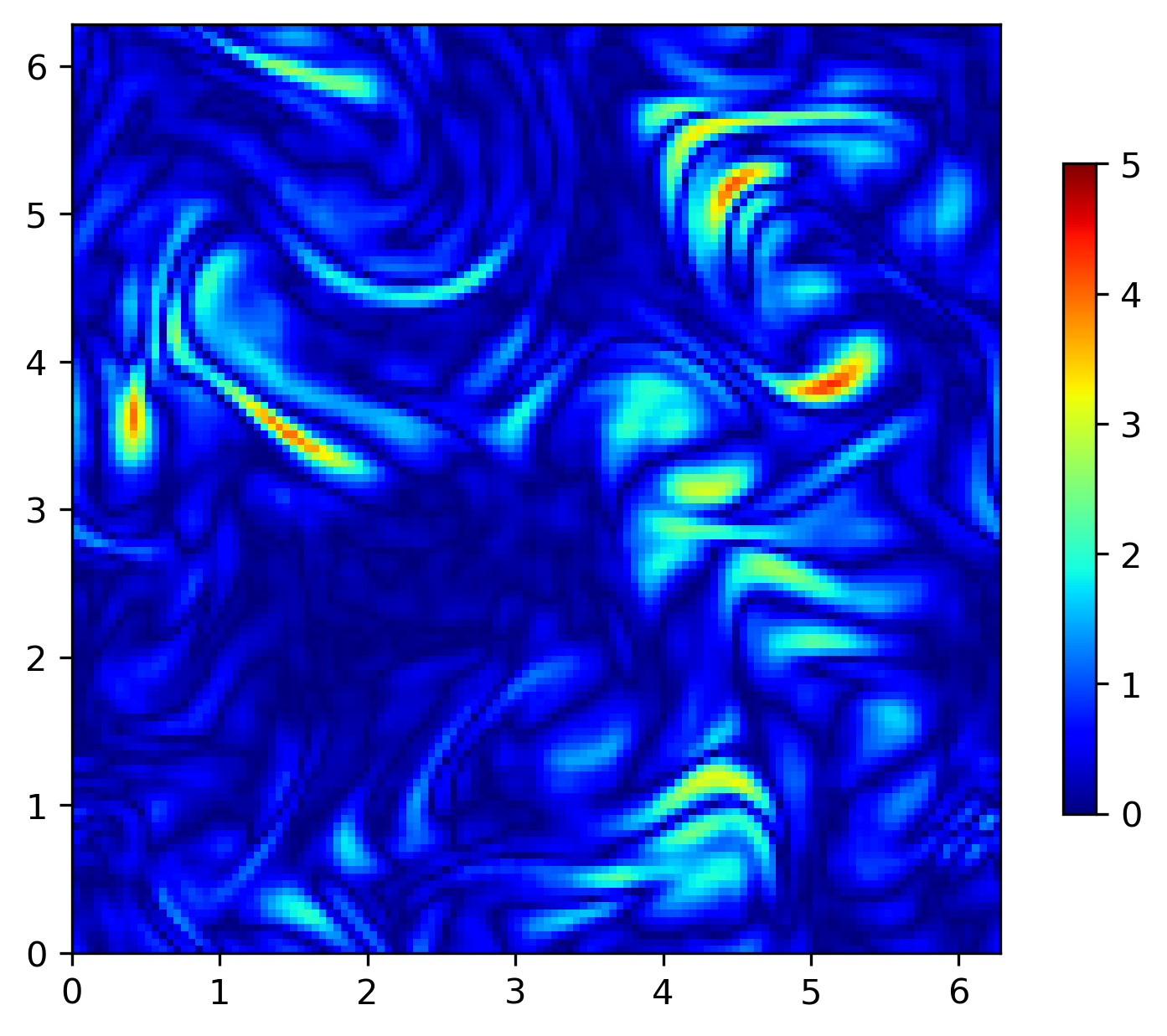} &
		\includegraphics[width=\linewidth]{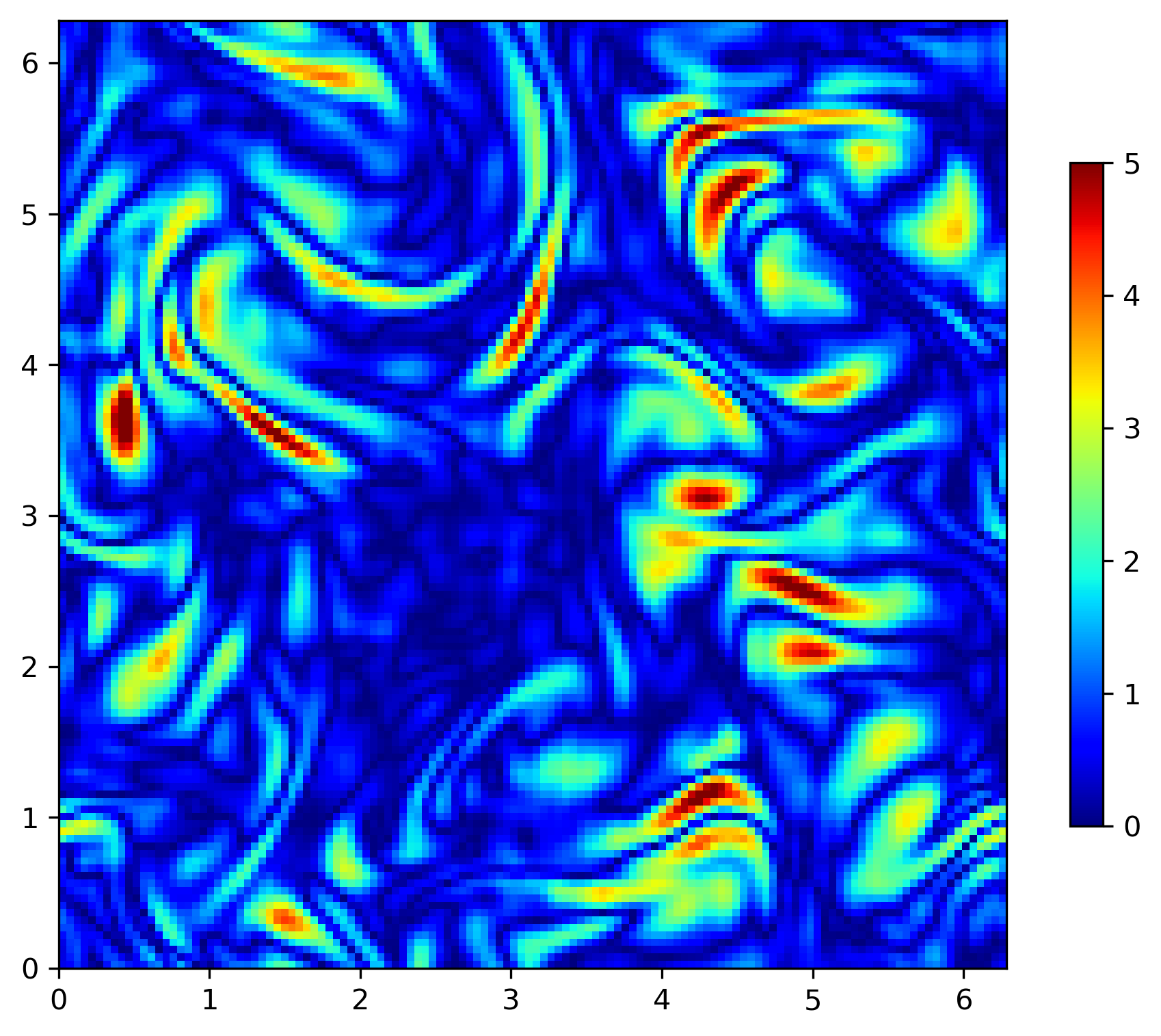} &
		\includegraphics[width=\linewidth]{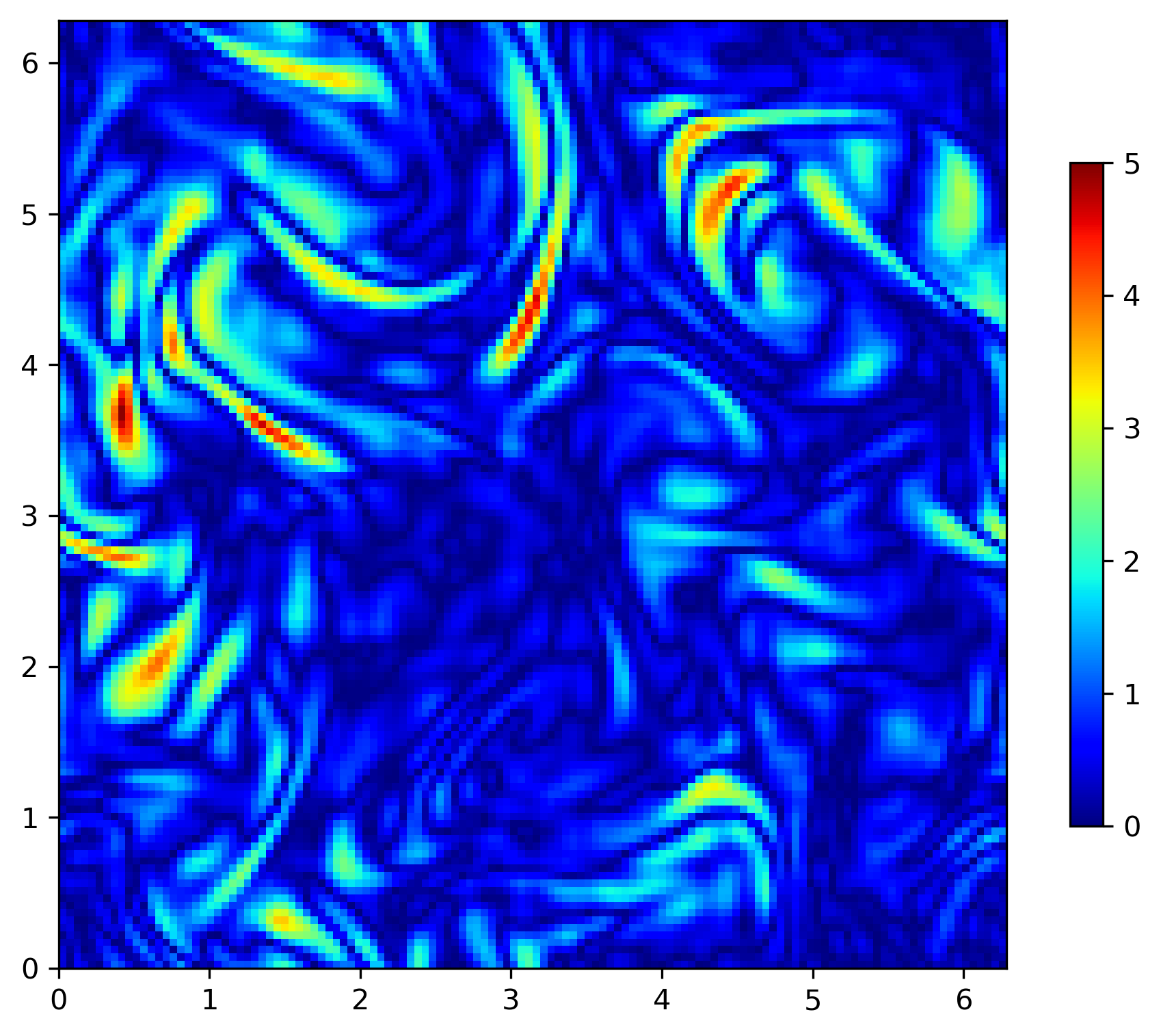} &
		\includegraphics[width=\linewidth]{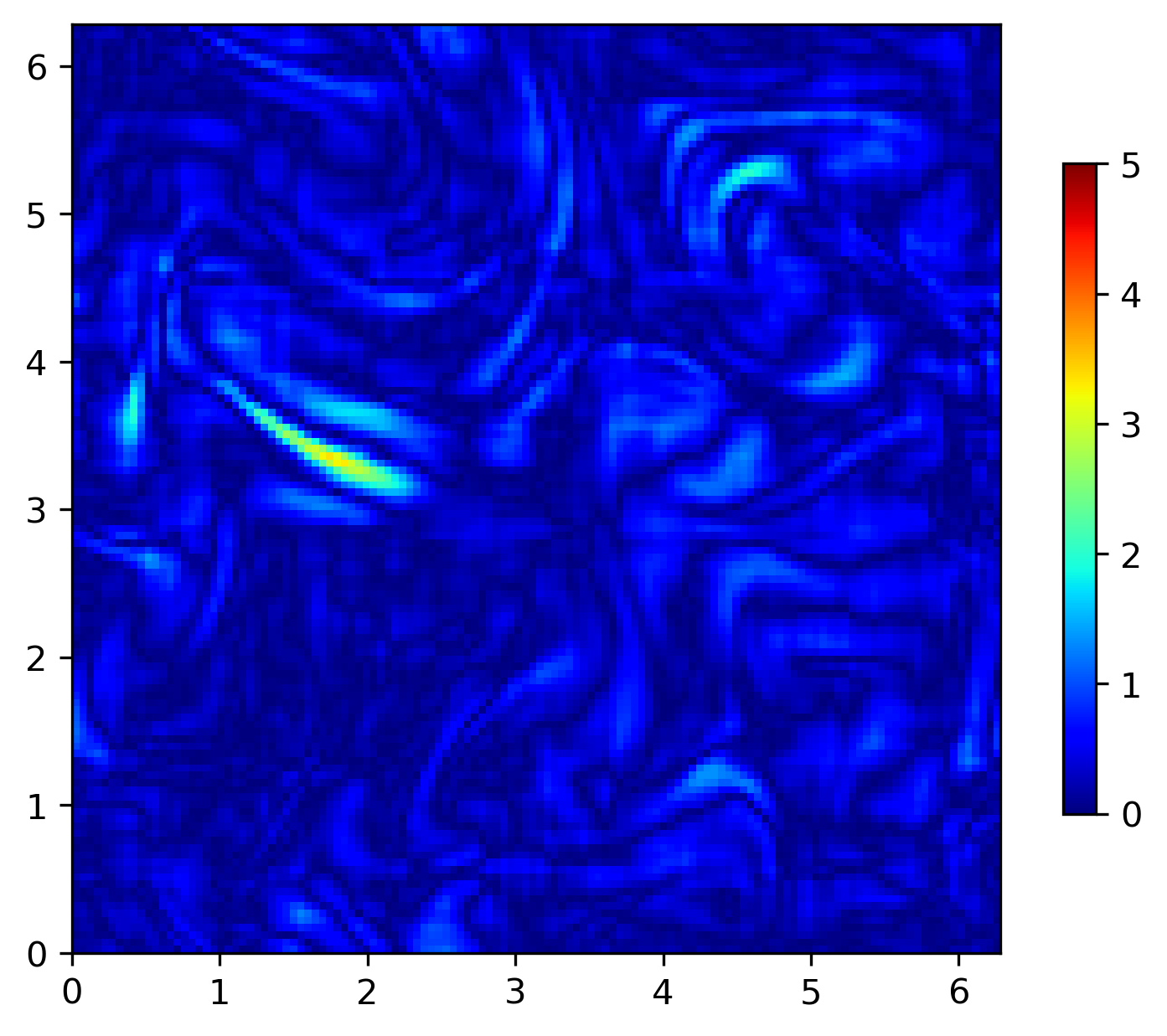} \\
	\end{tabular}
	\caption{Visual comparison of predicted vorticity field and absolute errors for (2+1)D Navier-Stokes equation at $t=1$.} \label{fig:navier-stokes}
\end{figure}

Table \ref{tab:navier-stokes} quantitatively compares the prediction accuracy of SPINN, TT-PINN, Tucker-PINN and HRE-PINN on the challenging (2+1)D decaying turbulence Navier-Stokes benchmark, reporting RMSE, relative $L^2$ error and relative $L^\infty$ error for vorticity fields. This chaotic viscous flow problem with strong nonlinear advection terms poses the strictest test among all numerical cases. HRE-PINN achieves the minimal error across all three evaluation metrics, yielding an RMSE of $0.414\,919$, a relative $L^2$ error of $0.073\,403$ and a relative $L^\infty$ error of $0.159\,532$. Compared with the best-performing baseline SPINN, HRE-PINN reduces the errors by approximately $40\%$. TT-PINN and Tucker-PINN produce substantially larger errors due to their fixed, inflexible structures, which fail to fully capture the intricate long-range vortex correlations in turbulent flow fields.
Figure \ref{fig:navier-stokes} visualizes the predicted vorticity fields and corresponding absolute errors of all competing models at $t=1$. SPINN, TT-PINN and Tucker-PINN exhibit prominent clustered error hotspots; these widespread residual artifacts demonstrate that tensor decompositions with fixed underlying structures lack sufficient expressive capacity to resolve fine turbulent flow structures. In contrast, the error of HRE-PINN remains uniformly suppressed at low magnitude over the entire periodic domain, with only negligible residual values visible near small-scale vortices.
This qualitative observation is fully consistent with the quantitative metrics in Table \ref{tab:navier-stokes}, and validates that the hierarchical framework equipped with automatic rank-evolving mechanism can accurately resolve complex chaotic fluid dynamics without manually presetting tensor ranks. For viscous turbulence problems, the rank-evolving mechanism automatically reveals the underlying low-rank structure of vorticity fields, yielding high-fidelity flow predictions that outperform existing tensor-based PINNs.

\section{Discussion}

In this section, we conduct a series of targeted ablation studies. We separately quantify the performance contributions of the hierarchical design, the rank-evolving mechanism, and the flexible underlying structure. Through controlled experiments on the representative 5D Poisson equation, we isolate the gain brought by each key component, and clarify why the proposed HRE representation outperforms existing tensor-based PINNs.

\subsection{Contribution of hierarchical design}

To demonstrate the advantages of our hierarchical design, we conduct ablation experiments on the 5D Poisson equation by comparing three configurations: 1. Inner representation only: the target solution function is represented by the continuous variant of SVDinsTN decomposition; 2. Outer representation only: the target solution function is represented by the rank-evolving variant of function Tucker decomposition; 3. Both: the target solution function is represented by the proposed HRE representation.

\begin{table}[width=.8\linewidth,cols=4,pos=!htb]
	\caption{Performance comparison for ablation study of hierarchical design on 5D Poisson equation} \label{tab:hierarchical}
	\begin{tabular*}{\tblwidth}{@{}LLLLL@{}}
		\toprule
		Inner representation & Outer representation & RMSE & $L^2$ error & $L^\infty$ error \\
		\midrule
		\ding{51} & \ding{55} & 0.001\,817 & 0.000\,561 & 0.003\,104 \\
		\ding{55} & \ding{51} & 0.001\,957 & 0.000\,604 & 0.002\,921 \\
		\ding{51} & \ding{51} & 0.001\,078 & 0.000\,333 & 0.001\,675 \\
		\bottomrule
	\end{tabular*}
\end{table}

Table \ref{tab:hierarchical} quantifies the individual and combined performance gains brought by the hierarchical architecture of HRE on the 5D Poisson equation. When only the inner representation is utilized, the model attains an RMSE of $0.001\,817$, a relative $L^2$ error of $0.000\,561$ and a relative $L^\infty$ error of $0.003\,104$, suffering from limited efficiency in decomposing high-dimensional coordinates. If we retain only the outer representation, the error rises to RMSE at $0.001\,957$, relative $L^2$ error at $0.000\,604$ and relative $L^\infty$ at $0.002\,921$, as the core tensor cannot directly capture complex cross-mode couplings. In contrast, the complete HRE hierarchical design integrating both inner and outer layers achieves the lowest RMSE ($0.001\,078$), relative $L^2$ error ($0.000\,333$) and relative $L^\infty$ error ($0.001\,675$), nearly halving the error magnitude.
This ablation result clearly verifies the complementary effect of the two hierarchical components: the outer representation mitigates the curse of dimensionality, while the inner representation accurately encodes multi-dimensional correlations; only their joint deployment delivers optimal approximation performance for high-dimensional PDE solutions.

\subsection{Contribution of rank-evolving mechanism}

The rank-evolving mechanism serves as the core adaptive component of the HRE representation, which automatically determines the suitable ranks and underlying structure throughout training. To evaluate the effectiveness of rank-evolving mechanism, we conduct ablation experiments on the 5D Poisson equation by removing part/all of the rank-evolving vectors in HRE representation. All other settings and architectures are kept identical.

\begin{table}[width=.8\linewidth,cols=4,pos=!htb]
	\caption{Performance comparison for ablation study of rank-evolving mechanism on 5D Poisson equation} \label{tab:rank evolving}
	\begin{tabular*}{\tblwidth}{@{}LLLLL@{}}
		\toprule
		Inner rank-evolving & Outer rank-evolving & RMSE & $L^2$ error & $L^\infty$ error \\
		\midrule
		\ding{55} & \ding{55} & 0.001\,816 & 0.000\,561 & 0.004\,377 \\
		\ding{51} & \ding{55} & 0.001\,463 & 0.000\,452 & 0.002\,636 \\
		\ding{55} & \ding{51} & 0.001\,397 & 0.000\,431 & 0.001\,823 \\
		\ding{51} & \ding{51} & 0.001\,078 & 0.000\,333 & 0.001\,675 \\
		\bottomrule
	\end{tabular*}
\end{table}

Table \ref{tab:rank evolving} clearly quantifies the independent and joint contributions of inner and outer rank-evolving mechanisms. Disabling both rank-evolving mechanisms yields the largest error metrics (RMSE of $0.001\,816$, relative $L^2$ error of $0.000\,561$, relative $L^\infty$ error of $0.004\,377$), which reveals the performance bottleneck of fixed manually preassigned ranks. Activating merely the inner rank-evolving vectors cuts the relative $L^2$ error to $0.000\,452$, while only enabling the outer rank-evolving vectors reduces the relative $L^2$ error to $0.000\,431$. When both inner and outer rank-evolving mechanisms are activated, the model reaches the minimal RMSE of $0.001\,078$, relative $L^2$ error of $0.000\,333$, and relative $L^\infty$ error of $0.001\,675$. This observation confirms that the two sparse rank-evolving vectors act cooperatively, and their synergy substantially improves the approximation precision for high-dimensional PDE solutions.

\subsection{Influence of underlying structure}

To quantitatively verify the influence of the underlying structure on model performance, we conduct ablation experiments on the 5D Poisson equation. In these experiments, the outer rank-evolving linear projection is kept unchanged, and only the inner decomposition is replaced with four fixed structures: tensor directly, TT decomposition, TR decomposition, and FCTN decomposition.

\begin{table}[width=.8\linewidth,cols=3,pos=!htb]
	\caption{Performance comparison of different underlying structures on 5D Poisson equation} \label{tab:underlying structure}
	\begin{tabular*}{\tblwidth}{@{}LLLL@{}}
		\toprule
		Underlying structure & RMSE & $L^2$ error & $L^\infty$ error \\
		\midrule
		Tensor directly & 0.001\,890 & 0.000\,583 & 0.003\,263 \\
		TT decomposition & 0.001\,772 & 0.000\,547 & 0.004\,088 \\
		TR decomposition & 0.001\,693 & 0.000\,523 & 0.003\,337 \\
		FCTN decomposition & 0.001\,397 & 0.000\,431 & 0.001\,823 \\
		Proposed & 0.001\,078 & 0.000\,333 & 0.001\,675 \\
		\bottomrule
	\end{tabular*}
\end{table}

Table \ref{tab:underlying structure} quantitatively demonstrates the superiority of the customized tensor network decomposition adopted in HRE representation over several fixed tensor decomposition structures. Direct tensor representation yields the highest RMSE and relative $L^2$ error among all candidates, failing to compress the high-dimensional core tensor and incurring severe redundant parameters. TT decomposition and TR decomposition only model adjacent-mode connections, resulting in moderate approximation accuracy but limited capacity to capture global cross-dimensional correlations. FCTN decomposition with fully connected structure achieves better fitting results yet still relies on fixed preset ranks. Proposed underlying structure equipped with rank-evolving vectors attains the minimal RMSE of $0.001\,078$, relative $L^2$ error of $0.000\,333$ and relative $L^\infty$ error of $0.001\,675$, as it can automatically simplify redundancy via sparsity regularization while retaining critical multi-mode interactions. This ablation validates that the customizable, rank-evolved underlying structure inside HRE representation provides stronger representation ability than conventional tensor decompositions with fixed structures for high-dimensional PDE solutions.

\section{Conclusion}

In this paper, we proposed a hierarchical rank-evolving (HRE) representation for multivariate functions and built HRE-PINNs to address two key drawbacks of existing tensor-based PINNs: pre-specified low-rank tensor decompositions with fixed underlying structures and manually tuned ranks.
The HRE representation adopts a hierarchical architecture. Its outer representation reduces the computational complexity of function evaluation from exponential to linear with respect to the dimensions, effectively alleviating the curse of dimensionality. The inner representation leverages a customized tensor network decomposition to capture the underlying structure of the target multivariate function. Equipped with learnable rank-evolving vectors and sparse regularization, our method automatically determines the suitable ranks and underlying structure during training, removing tedious manual hyperparameter tuning.
Extensive experiments on various PDE benchmarks, including high-dimensional static problems (3D Helmholtz equation and 5D Poisson equation), nonlinear time-dependent problems ((2+1)D Klein-Gordon equation), and complex fluid-dynamics problems ((2+1)D flow mixing equation and (2+1)D Navier-Stokes equation), show HRE-PINN outperforms existing tensor-based PINNs across all error metrics. Ablation studies verify the indispensable performance gains brought by both the hierarchical design and rank-evolving mechanism.

In summary, the HRE representation provides a novel function approximation paradigm. The resulting HRE-PINN offers an accurate, easy-to-tune alternative to existing tensor-based PINNs, with broad application prospects for high-dimensional PDE simulation across fluid dynamics, quantum mechanics, and engineering multi-physics modeling.




%
%
%
%






\bibliographystyle{elsarticle-num-names}

\bibliography{./refs}



\end{document}